\documentclass[11pt]{article}

\usepackage[margin=1in]{geometry}
\usepackage[utf8]{inputenc}
\usepackage[T1]{fontenc}
\usepackage{lmodern}
\usepackage{microtype}
\usepackage{graphicx}
\usepackage{amsmath,amssymb,bm}
\usepackage{booktabs}
\usepackage{array}
\usepackage{multirow}
\usepackage{longtable}
\usepackage{tabularx}
\usepackage{siunitx}
\usepackage{xcolor}
\usepackage{algorithm}
\usepackage{algpseudocode}
\usepackage{enumitem}
\usepackage{caption}
\usepackage{subcaption}
\usepackage{url}
\usepackage[colorlinks=true,linkcolor=blue,citecolor=blue,urlcolor=blue]{hyperref}
\usepackage[nameinlink,noabbrev]{cleveref}

\setlist[itemize]{topsep=2pt,itemsep=2pt,parsep=0pt,partopsep=0pt}
\setlist[enumerate]{topsep=2pt,itemsep=2pt,parsep=0pt,partopsep=0pt}
\newcommand{\method}{AtteConDA}
\newcommand{\pam}{Patch-wise Adaptation Module}
\newcommand{\pamshort}{PAM}
\newcommand{\etal}{\textit{et al.}}
\newcommand{\codeurl}{\url{https://github.com/ShogoNoguchi/AtteConDA}}

\title{\method: Attention-Based Conflict Suppression in Multi-Condition Diffusion Models and Synthetic Data Augmentation}
\author{Shogo Noguchi\\Gunma University}
\date{2026}

\begin{document}
\maketitle

\begin{abstract}
Controllable image generation has made it possible to synthesize images from structural conditions such as sketches, human poses, semantic segmentation maps, depth maps, and edges. This capability is attractive for data augmentation because generated images may preserve existing annotations while changing appearance. For high-level autonomous-driving tasks, however, the required annotations are not limited to class masks or bounding boxes. They may include traffic-rule structure, lane-to-sign relationships, map topology, driving explanations, and graph-based visual question answering. Such annotations are expensive, context-dependent, and difficult to re-create manually. A useful generative augmentation method must therefore change weather and time-of-day appearance while preserving detailed road-scene structure.

This paper presents \method, a synthetic data augmentation framework based on multi-condition diffusion models and explicit condition-conflict suppression. The generator uses semantic segmentation, relative depth, and edge maps extracted from the original image, thereby constraining class layout, geometric structure, and local contours at the same time. To reuse strong controllable-generation priors, the local control branch is initialized from Uni-ControlNet-compatible pretrained weights. To address the conflict that arises when multiple local conditions impose incompatible guidance at the same spatial location, \method introduces the \pam{} (\pamshort). \pamshort{} uses a tri-attention-style score computation over semantic, depth, and edge features and applies straight-through hard selection so that the locally effective condition can be selected while remaining trainable.

The framework also includes prompt generation, inference, and evaluation protocols for structure-preserving augmentation in autonomous-driving scenes. Experiments use a combined training set of 146,809 driving-related images and evaluate on 3,048 Waymo front-camera images. The results show that Uni-ControlNet initialization substantially improves structure preservation compared with scratch training, that longer fine-tuning improves most structural and quality metrics, and that \pamshort{} improves all four structural metrics at 60K steps. In comparison with DGInStyle, the proposed model is weaker in semantic mIoU but stronger in depth consistency, edge consistency, object preservation, and realism. These results indicate that high-level annotation-preserving augmentation requires more than semantic-mask fidelity: it must preserve geometry, contours, object existence, and image realism jointly.
\end{abstract}

\textbf{Code availability.} The implementation, project page source, model release information, and reproducibility documentation are available at \codeurl. The paper intentionally omits the thesis acknowledgements for arXiv-style submission.

\tableofcontents

\section{Introduction}
\label{sec:introduction}

\subsection{From image recognition to high-level visual understanding}

Recent image AI has rapidly expanded from low-level recognition centered on image classification to object detection, semantic segmentation, instance-level understanding, vision-language explanation, reasoning, and decision making. Large-scale ImageNet classification established the practicality of deep visual representations, and subsequent advances such as ResNet, FCN, U-Net, DeepLab-style dense predictors, and DETR-style end-to-end detectors greatly improved urban-scene perception \cite{krizhevsky2012alexnet,he2016resnet,long2015fcn,ronneberger2015unet,chen2018deeplabv3plus,carion2020detr}. At the same time, CLIP, BLIP-2, and LLaVA made it increasingly natural to describe image content in language and to condition visual models on text \cite{radford2021clip,li2023blip2,liu2023llava}. In autonomous driving, the emphasis has also shifted from recognition accuracy alone to explanations of vehicle behavior, hazardous-object reasoning, interactive assistance, and language-mediated high-level understanding, as seen in Talk2Car, DriveLM, DriveGPT4, and LanguageMPC \cite{deruyttere2019talk2car,sima2024drivelm,xu2023drivegpt4,sha2023languagempc}.

This shift increases both the number and the cost of annotations attached to each image. Earlier datasets mainly required class labels, bounding boxes, or pixel labels. In contrast, current high-level tasks require labels such as which object is dangerous, why a driving action should be chosen, which sign or traffic light governs which lane, or which object should be referred to during lane change. Such labels are context-dependent and require semantic understanding, so they usually require human annotation. This creates a bottleneck for large-scale dataset construction. A data augmentation system that can reuse existing annotations while generating diverse visual conditions is therefore important.

\subsection{Large autonomous-driving datasets and the scarcity of high-level data}

Autonomous-driving datasets have grown rapidly over the last decade. KITTI provided an early vehicle-mounted multimodal benchmark with 14,999 images; Cityscapes provided 5,000 fine annotations and 20,000 coarse annotations; and Mapillary Vistas collected 25,000 high-resolution street-view images across diverse global conditions \cite{geiger2012kitti,cordts2016cityscapes,neuhold2017mapillary}. BDD100K expanded the field to 100,000 videos and ten tasks; nuScenes provided 1,000 scenes, 40,000 keyframes, and 1.4M images; and Waymo Open Dataset and Argoverse 2 pushed the scale of real-driving perception, prediction, and map benchmarks further \cite{yu2020bdd100k,caesar2020nuscenes,sun2020waymo,wilson2023argoverse2}.

However, the center of this large-scale trend is still low- and mid-level perception: detection, tracking, segmentation, and 3D box estimation. The high-level tasks targeted by this work, such as traffic-rule extraction, lane-to-sign correspondence, HD-map topology reasoning, traffic knowledge graph generation, and graph visual question answering, are only recently becoming available as public benchmarks \cite{wang2023openlanev2,guo2023vtkgg,chang2025mapdr,sima2024drivelm,wei2025driveqa}. \Cref{tab:dataset_scale_contrast} summarizes the difference. Low-level perception benchmarks can reach 100K-video or million-image scale, while high-level benchmarks containing traffic rules, graph relationships, map topology, or reasoning labels often remain at thousands to tens of thousands of samples. This difference is natural because high-level labels require interpreting sign text, assigning signs to lanes, structuring traffic rules, writing explanations, or constructing question-answer pairs. Consequently, a generative method that preserves the original structure and changes only appearance has particular value for high-level tasks.

\begin{table}[tbp]
\centering
\caption{Scale comparison between large low-level autonomous-driving benchmarks and high-level benchmarks. The numbers are reported as dataset-level descriptions and are used here to motivate annotation-preserving augmentation.}
\label{tab:dataset_scale_contrast}
\resizebox{\textwidth}{!}{%
\begin{tabular}{p{18mm}p{32mm}p{44mm}p{48mm}p{70mm}}
\toprule
Category & Dataset & Main task & Scale & Note \\
\midrule
Low-level & KITTI \cite{geiger2012kitti} & 2D/3D detection and geometry benchmark & 14,999 images (7,481 train + 7,518 test) & Early standard vehicle-mounted multimodal benchmark. \\
Low-level & Cityscapes \cite{cordts2016cityscapes} & Semantic and instance segmentation & 5,000 fine + 20,000 coarse images & High-quality pixel labels for urban scenes. \\
Low-level & Mapillary Vistas \cite{neuhold2017mapillary} & Semantic understanding and instance-aware labeling & 25,000 images & Street images with geographic, weather, and illumination diversity. \\
Low-level & BDD100K \cite{yu2020bdd100k} & Detection, lane, drivable area, MOT, segmentation, and other tasks & 100,000 videos & Heterogeneous multitask benchmark; many image tasks annotate a reference frame per video. \\
Low-level & nuScenes \cite{caesar2020nuscenes} & Multisensor perception, 3D detection, and tracking & 1,000 scenes, 40,000 keyframes, 1.4M images & Representative benchmark with 23 classes and a 360-degree sensor suite. \\
High-level / real & OpenLane-V2 \cite{wang2023openlanev2} & 3D HD-map topology reasoning & 2,000 annotated road scenes & Adds lane and traffic-element relationships to existing data. \\
High-level / real & RS10K \cite{guo2023vtkgg} & Traffic knowledge graph generation & 10,066 images and 104,428 relations & Annotates relationships among roads, lanes, signs, and scene elements. \\
High-level / real & MapDR \cite{chang2025mapdr} & Traffic-rule extraction and rule-lane correspondence & More than 10,000 clips, 400,000 images, and more than 18,000 rules & Extracts lane-level rules from traffic signs and associates them with vectorized HD maps. \\
High-level / real & DriveLM-nuScenes \cite{sima2024drivelm,wei2025driveqa} & Graph VQA and driving reasoning & 4,871 images and 443K QA pairs & Real-world subset; reasoning labels are high-level but real-world image scale is limited. \\
High-level / synthetic-mixed & DriveLM-CARLA \cite{sima2024drivelm,wei2025driveqa} & Graph VQA and driving reasoning & 64,285 images and 1.566M QA pairs & The larger part of DriveLM uses CARLA simulation. \\
High-level / synthetic-mixed & DriveQA-T / DriveQA-V \cite{wei2025driveqa} & Traffic-rule QA and right-of-way reasoning & 26K text QA, 68K images, 448K QA (474K samples total) & Combines CARLA procedural generation with Mapillary real images to cover long-tail traffic rules. \\
\bottomrule
\end{tabular}}
\end{table}

\subsection{Conventional augmentation and its limitations}

Data augmentation is a classical and powerful technique in deep learning. Surveys consistently report the importance of augmentation for generalization \cite{shorten2019survey,yang2022imageaug,kumar2023dataaugapproaches,wang2024datasurvey,mumuni2024syntheticdatasurvey}. Classical transformations such as flipping, rotation, translation, random cropping, and color transformation, as well as stronger methods such as Cutout, mixup, AutoAugment, RandAugment, AugMix, TrivialAugment, and CutMix, are effective for classification, detection, and robustness \cite{devries2017cutout,zhang2018mixup,cubuk2019autoaugment,cubuk2020randaugment,hendrycks2020augmix,muller2021trivialaugment,yun2019cutmix}.

\begin{figure}[tbp]
\centering
\includegraphics[width=0.95\linewidth]{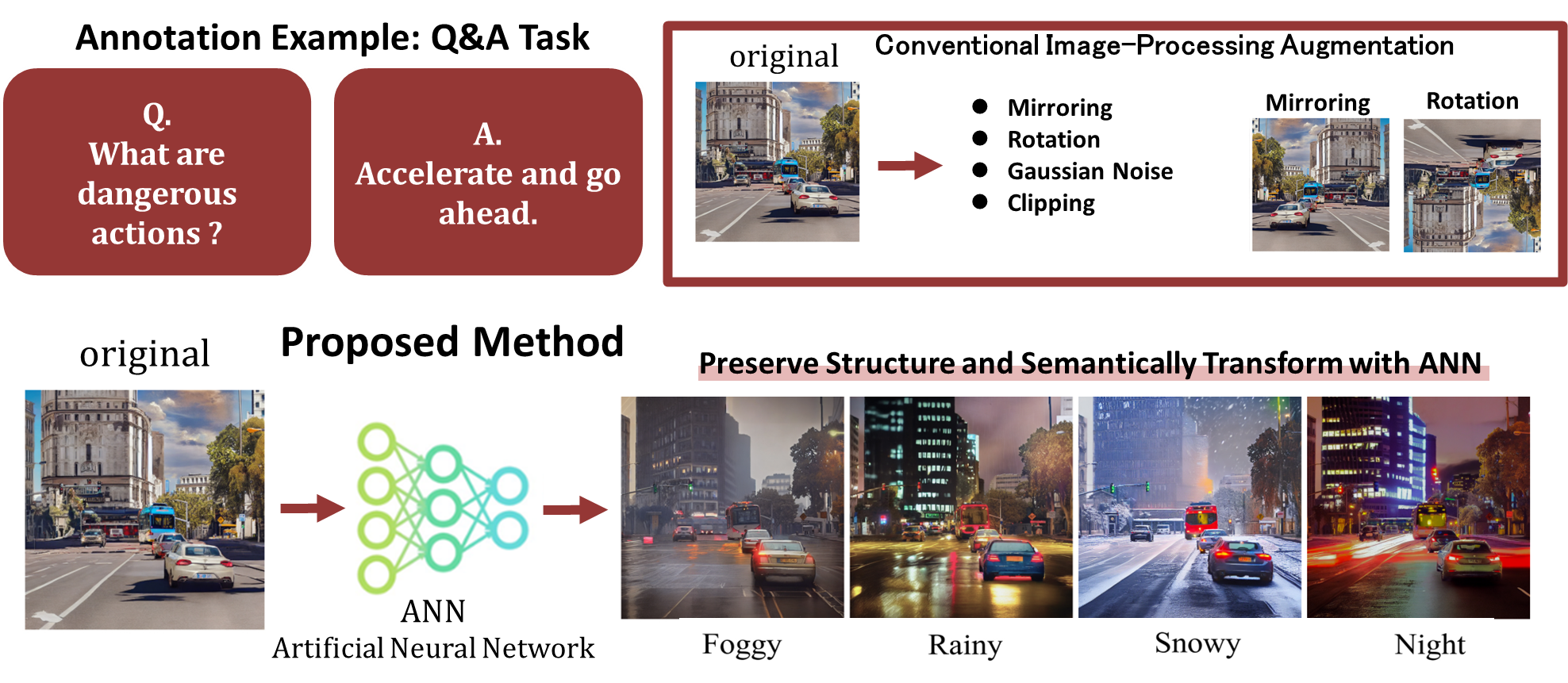}
\caption{Conceptual comparison between conventional image-engineering augmentation and generative augmentation. Conventional operations such as mirroring, rotation, noise, and clipping increase sample count but do not create semantic changes in weather or time. Generative augmentation can change appearance conditions such as fog, rain, snow, and night while preserving the original scene structure.}
\label{fig:augmentation_comparison}
\end{figure}

Nevertheless, most conventional augmentations depend on local pixel changes or statistical perturbations of existing images. They do not generate semantic changes in appearance conditions such as rain, snow, night illumination, road-surface reflection, fog, or backlight. In autonomous driving, such environmental shifts are often a major source of performance degradation; conventional augmentation alone is therefore insufficient to guarantee domain-shift robustness \cite{yang2022imageaug,mumuni2024syntheticdatasurvey,yu2020bdd100k,caesar2020nuscenes,sun2020waymo}. This work does not reject conventional augmentation. Rather, it treats generative models as a way to supply the semantic diversity that conventional augmentation cannot provide. \Cref{fig:augmentation_comparison} illustrates this distinction. If the original image structure is preserved, the annotations attached to that image may also remain reusable.

\subsection{Synthetic data and generative augmentation}
\label{sec:gen_aug}

Autonomous-driving research has long used synthetic data and simulation to reduce the cost of real-world capture and manual labeling. SYNTHIA, GTA5 / Playing for Data, CARLA, and Synscapes construct virtual urban environments and automatically generate pixel labels and geometry \cite{ros2016synthia,richter2016playingfordata,dosovitskiy2017carla,wrenninge2018synscapes}. Domain randomization also organizes the idea of randomizing appearance to improve real-world generalization \cite{tremblay2018domainrandomization}. These approaches are important, but they are not directly designed for the use case of taking one real image as input, preserving its structure, and changing only its appearance.

From the perspective of image-to-image translation, GAN-based methods created an earlier line of work. GANs established a foundation for high-quality generation \cite{goodfellow2014gan}; pix2pix generalized paired image-to-image translation; CycleGAN generalized unpaired domain translation; and UNIT, MUNIT, and SPADE developed shared-domain representations, multimodality, and semantic-conditioned synthesis \cite{isola2017pix2pix,zhu2017cyclegan,liu2017unit,huang2018munit,park2019spade}. More recently, diffusion-based image editing and image translation have become dominant. Palette, SDEdit, Prompt-to-Prompt, and InstructPix2Pix enabled higher-quality editing \cite{saharia2022palette,meng2022sdedit,hertz2023prompttoprompt,brooks2023instructpix2pix}. ControlNet, T2I-Adapter, GLIGEN, and IP-Adapter then supplied mechanisms for adding external controls to large pretrained text-to-image diffusion models \cite{zhang2023controlnet,mou2024t2iadapter,li2023gligen,ye2023ipadapter}.

The most directly related line is annotation-conditioned data augmentation. DGInStyle generates pixel-aligned image-label pairs from a semantic mask and a style prompt, showing that diffusion-based generation can improve semantic segmentation in autonomous driving \cite{jia2024dginstyle}. SynDiff-AD similarly combines semantic-map ControlNet with subgroup-specific captions to generate images for underrepresented weather and time conditions, reporting improvements in semantic segmentation and end-to-end autonomous driving \cite{goel2024syndiff}. A survey by Li \etal{} emphasizes that the long-tail nature of autonomous-driving data makes it difficult to collect rare scenarios only from the real world, and that high-fidelity data generation and simulation are important for data-centric closed-loop development \cite{li2024datacentric}.

\begin{figure}[tbp]
\centering
\includegraphics[width=0.95\linewidth]{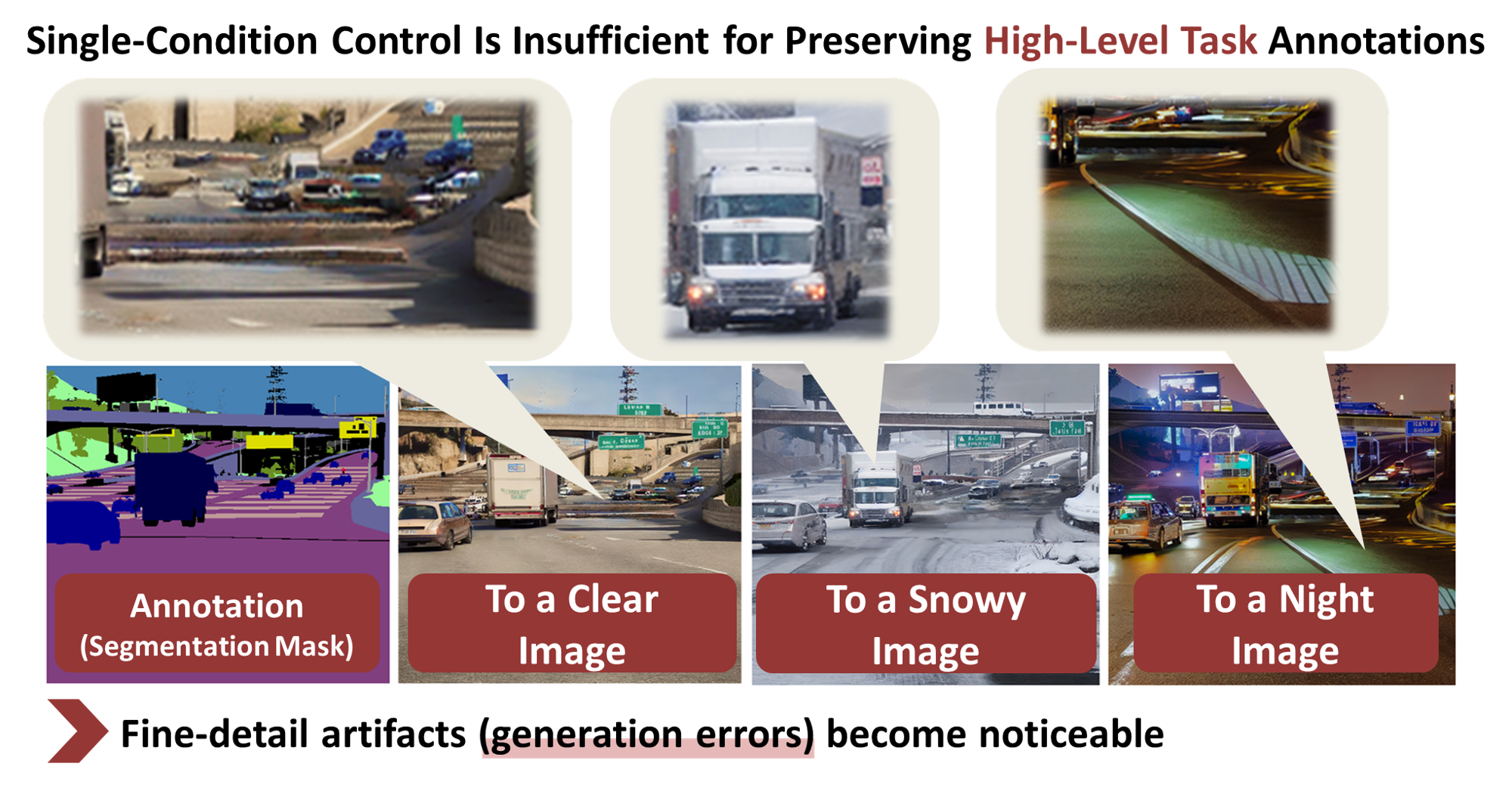}
\caption{Example of the limitation of semantic-mask-only generation, motivated by DGInStyle-style augmentation \cite{jia2024dginstyle}. If the semantic mask is the only structural condition, details not explicitly constrained by the mask can become weak, and artifacts can appear around road boundaries, vehicles, and other fine structures.}
\label{fig:dginstyle_failure}
\end{figure}

However, many direct predecessors mainly use semantic segmentation or semantic maps as structural conditions. Such masks preserve class regions but cannot fully constrain lane-marker continuity, road-marking shape, fine sign geometry, sign text, or relationships between signs and lanes. DGInStyle itself reports difficulties around small objects and artifacts during high-resolution generation, and semantic-only conditioning tends to leave regions not specified by the mask weakly constrained \cite{jia2024dginstyle}. SynDiff-AD also reports missing lane markers, distorted vehicle shapes, broken road markings, and insufficient signboard-text preservation \cite{goel2024syndiff}. \Cref{fig:dginstyle_failure} shows a representative conceptual failure. Thus, the core issue is not only synthesizing realistic images but also designing conditioning that does not destroy annotations required by higher-level downstream tasks.

\subsection{Multi-condition generation and condition conflict}

A single condition cannot express all constraints in an urban scene. A semantic map provides class layout, but lane-line details, depth-based geometry, local sign shape, and edge-level contours are different information. Therefore, composable and multi-conditional generation is an important research problem. Composer treats text, sketch, depth, palette, and other signals as composable conditions, but its supplementary material also shows a failure case in which text is underweighted when it conflicts with sketch and depth \cite{huang2023composer}. This demonstrates that simple addition of multiple conditions is insufficient; priority and consistency among conditions must be handled explicitly.

T2I-Adapter attaches lightweight adapters for diverse external conditions to pretrained diffusion models \cite{mou2024t2iadapter}. Uni-ControlNet handles local controls such as edge, depth, and segmentation masks and global controls such as CLIP image embeddings in an all-in-one framework \cite{zhao2023unicontrolnet}. AnyControl further introduces a unified multi-control framework for arbitrary combinations of control signals \cite{sun2024anycontrol}. Yet condition conflict remains a central problem. If different conditions overlap in the same region, one condition may cancel another, and local structure or semantics may collapse. PixelPonder explicitly points out that separate control branches in ControlNet-like methods can produce conflicting guidance during denoising, causing structural distortions and artifacts \cite{pan2025pixelponder}. It proposes patch-level adaptive condition selection, but a discrete selection mechanism must be made trainable, and the interaction among conditions should be modeled.

\begin{figure}[tbp]
\centering
\includegraphics[width=0.95\linewidth]{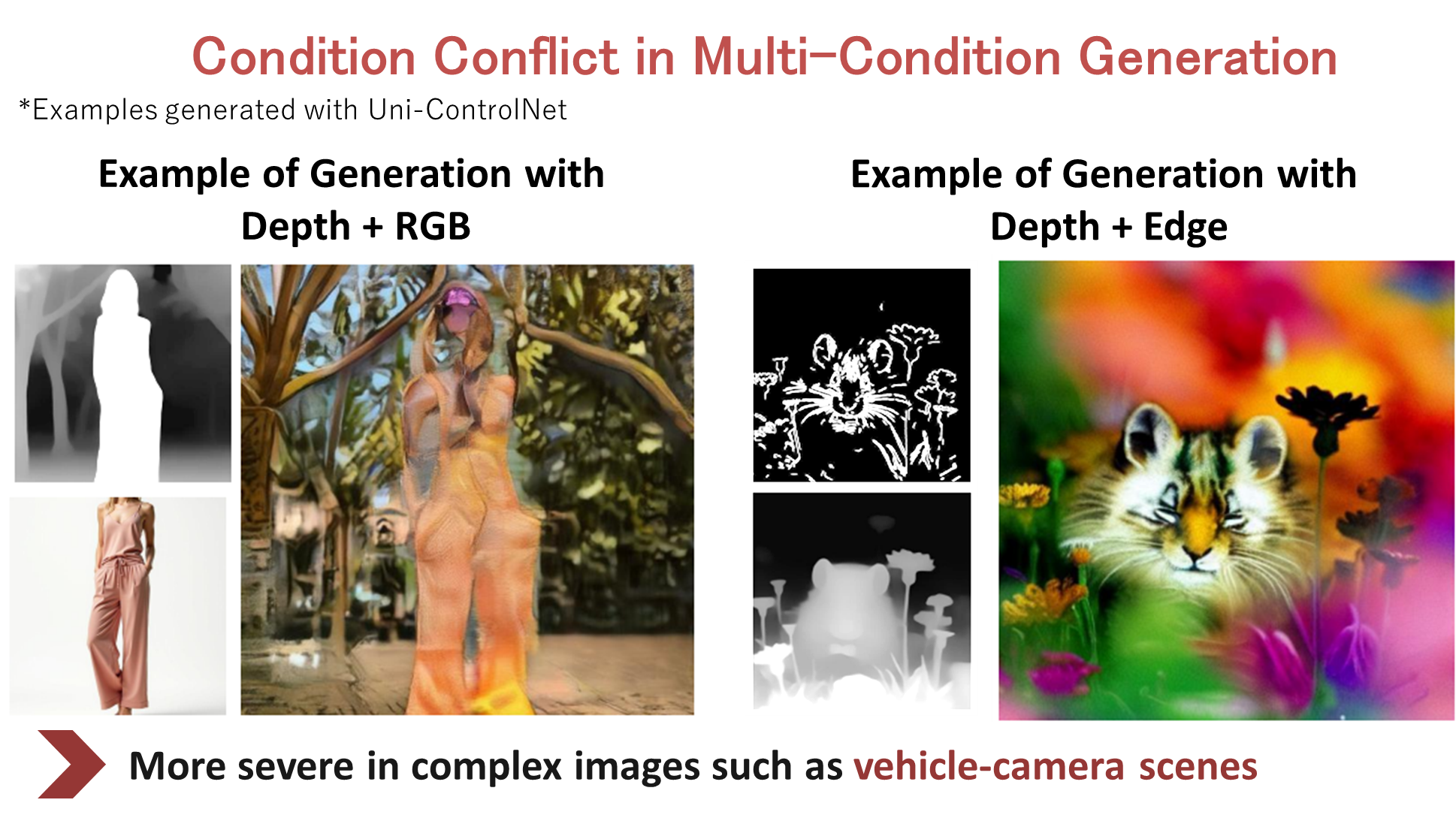}
\caption{Condition conflict in multi-condition generation. Depth provides low-frequency spatial and distance constraints, while RGB or edge constraints emphasize higher-frequency texture and contour information. When these conditions are injected together without conflict handling, model guidance can become inconsistent and produce distorted or semantically mismatched results.}
\label{fig:condition_conflict}
\end{figure}

\Cref{fig:condition_conflict} illustrates the conflict. Depth constrains low-frequency structure such as object placement and distance, whereas edges and RGB-like signals constrain high-frequency contours and textures. In driving scenes with trees, poles, distant road geometry, sign boundaries, and reflective surfaces, such conflicts become especially severe. This motivates a module that suppresses conflict rather than merely accumulating all conditions.

\subsection{Goal, problem statement, and contributions}

The goal of this work is to generate synthetic images that change only appearance conditions, such as weather and time of day, while preserving the annotations attached to the original image as much as possible. The target domain is autonomous driving because road-scene images require detailed preservation of cars, pedestrians, traffic lights, traffic signs, lanes, and road geometry, and because high-level tasks such as rule extraction and driving QA require expensive annotations.

This paper makes four main contributions.
\begin{itemize}
  \item It formulates high-level annotation-preserving augmentation as a multi-condition diffusion problem that uses semantic segmentation, depth, and edges rather than semantic masks alone.
  \item It shows that Uni-ControlNet-compatible pretrained local-control representations provide a useful initialization for autonomous-driving multi-condition generation.
  \item It introduces \pamshort, a conflict-suppression module that uses tri-attention-style condition scoring and straight-through hard selection to decide which condition should be used at each local feature position.
  \item It organizes a comparative evaluation framework covering structure preservation, realism, diversity, and text alignment, enabling comparison against prior methods and future models.
\end{itemize}

\section{Preliminaries}
\label{sec:preliminaries}

\subsection{Latent diffusion models and classifier-free guidance}

A variational autoencoder maps images from high-dimensional pixel space to a lower-dimensional latent space \cite{kingma2013aevb}. Latent Diffusion Models use this latent space to run the diffusion process more efficiently than pixel-space diffusion \cite{rombach2022ldm}. DDPM introduced the standard forward noise process and reverse denoising process for image generation \cite{sohl2015diffusion,ho2020ddpm}; improved denoising diffusion models refined variance learning and sampling design \cite{nichol2021improvedddpm}; and score-based generative modeling generalized diffusion through stochastic differential equations \cite{song2021sde}. DDIM provides deterministic non-Markovian sampling with fewer steps \cite{song2021ddim}. This work uses Stable-Diffusion-style latent diffusion because it provides a strong pretrained image prior and a text-conditioned generation mechanism. Classifier-free guidance combines conditional and unconditional predictions to strengthen text alignment without an external classifier \cite{ho2022cfg}.

\subsection{Controllable diffusion generation}

ControlNet freezes a pretrained text-to-image diffusion model and adds trainable control branches for spatial conditions such as edge, depth, and segmentation \cite{zhang2023controlnet}. T2I-Adapter adds lightweight adapters to extract controllability from pretrained models \cite{mou2024t2iadapter}. GLIGEN performs open-set grounding by injecting grounding inputs \cite{li2023gligen}. IP-Adapter integrates image prompts with text-compatible conditioning \cite{ye2023ipadapter}. MultiDiffusion fuses multiple generation paths for spatial control, and SDEdit, Prompt-to-Prompt, and InstructPix2Pix extend diffusion to editing \cite{bartal2023multidiffusion,meng2022sdedit,hertz2023prompttoprompt,brooks2023instructpix2pix}.

Uni-ControlNet is particularly relevant because it processes multiple local controls and global controls in one composable framework \cite{zhao2023unicontrolnet}. It supports Canny edge, MLSD edge, HED boundary, sketch, OpenPose, depth, and segmentation, and uses zero inputs for missing conditions. This compatibility is useful for the present work. The proposed base generator reuses Uni-ControlNet local-control representations, but uses only edge, depth, and semantic segmentation for the driving augmentation setting.

\subsection{Condition maps}

\textbf{Semantic segmentation.} Semantic segmentation assigns a class label to each pixel and captures scene layout. FCN, U-Net, DeepLabv3+, SegFormer, Mask2Former, and OneFormer form a major line of dense-prediction models \cite{long2015fcn,ronneberger2015unet,chen2018deeplabv3plus,xie2021segformer,cheng2022mask2former,jain2023oneformer}. This work uses a Cityscapes-oriented OneFormer model as the semantic-condition generator because it provides stable urban-scene structure.

\begin{figure}[tbp]
\centering
\includegraphics[width=0.92\linewidth]{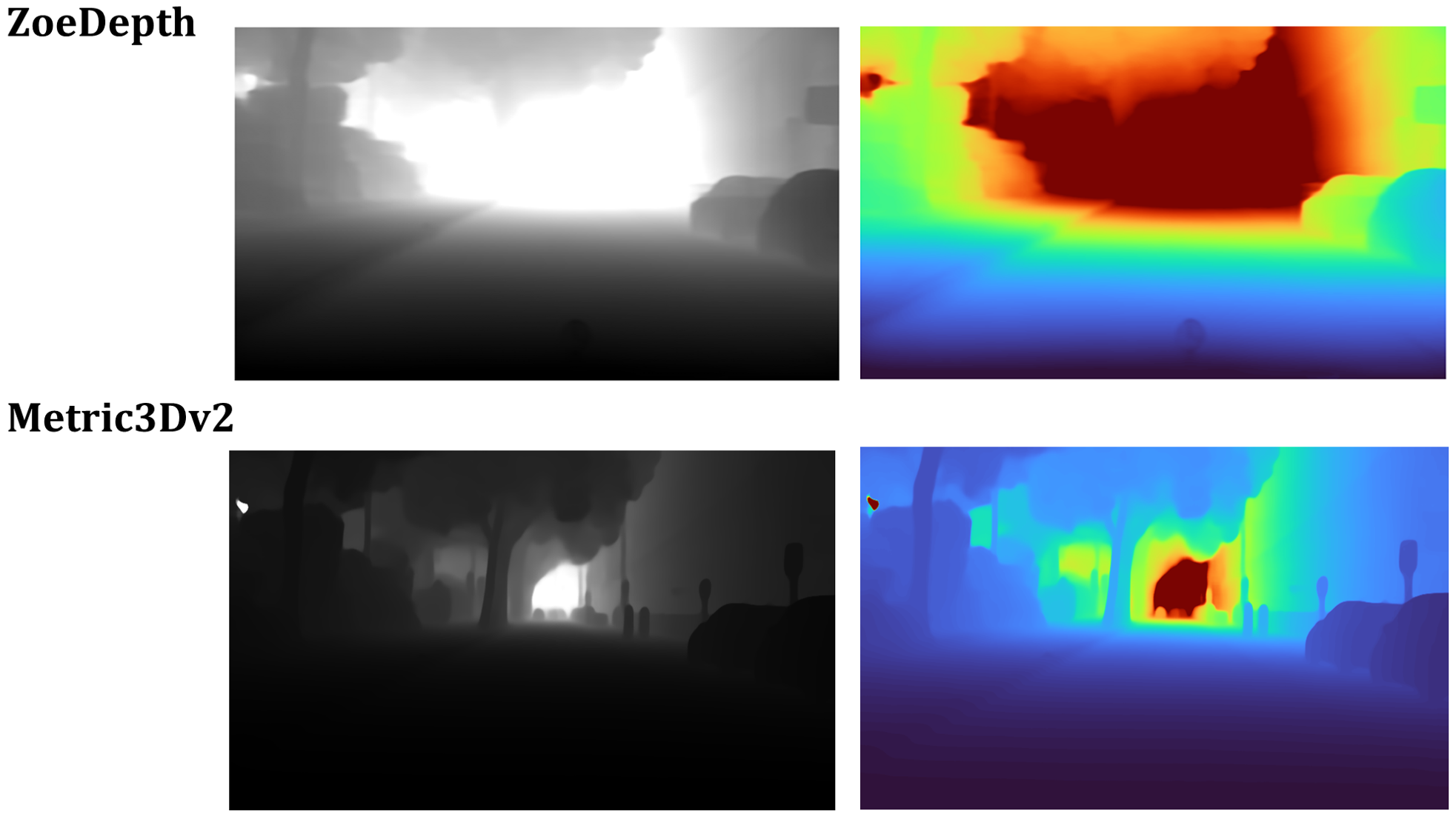}
\caption{Comparison of monocular depth maps. ZoeDepth can saturate in distant regions, whereas Metric3Dv2 preserves smoother distant gradients and clearer road-depth structure in this example. Since this work evaluates whether the generated image preserves the original perspective and relative geometry, stable relative depth is important.}
\label{fig:metric_depth_compare}
\end{figure}

\textbf{Depth.} Monocular depth estimation is important for 3D geometry in driving scenes. Eigen \etal{} proposed an early multiscale CNN approach \cite{eigen2014depth}; Monodepth2 improved self-supervised depth \cite{godard2019monodepth2}; DPT and MiDaS improved dense prediction and zero-shot transfer \cite{ranftl2021dpt,ranftl2022midas}; Depth Anything and Depth Anything V2 used large-scale unlabeled and synthetic data \cite{yang2024depthanything,yang2024depthanythingv2}; and ZoeDepth and Metric3Dv2 advanced relative and metric depth estimation \cite{bhat2023zoedepth,hu2024metric3dv2}. This work uses Metric3Dv2 as the depth projector. Because the camera intrinsics of all input images are not treated as known in the pipeline, the output is used as relative depth for geometry consistency rather than absolute metric distance. \Cref{fig:metric_depth_compare} compares representative depth maps.

\begin{figure}[tbp]
\centering
\includegraphics[width=0.90\linewidth]{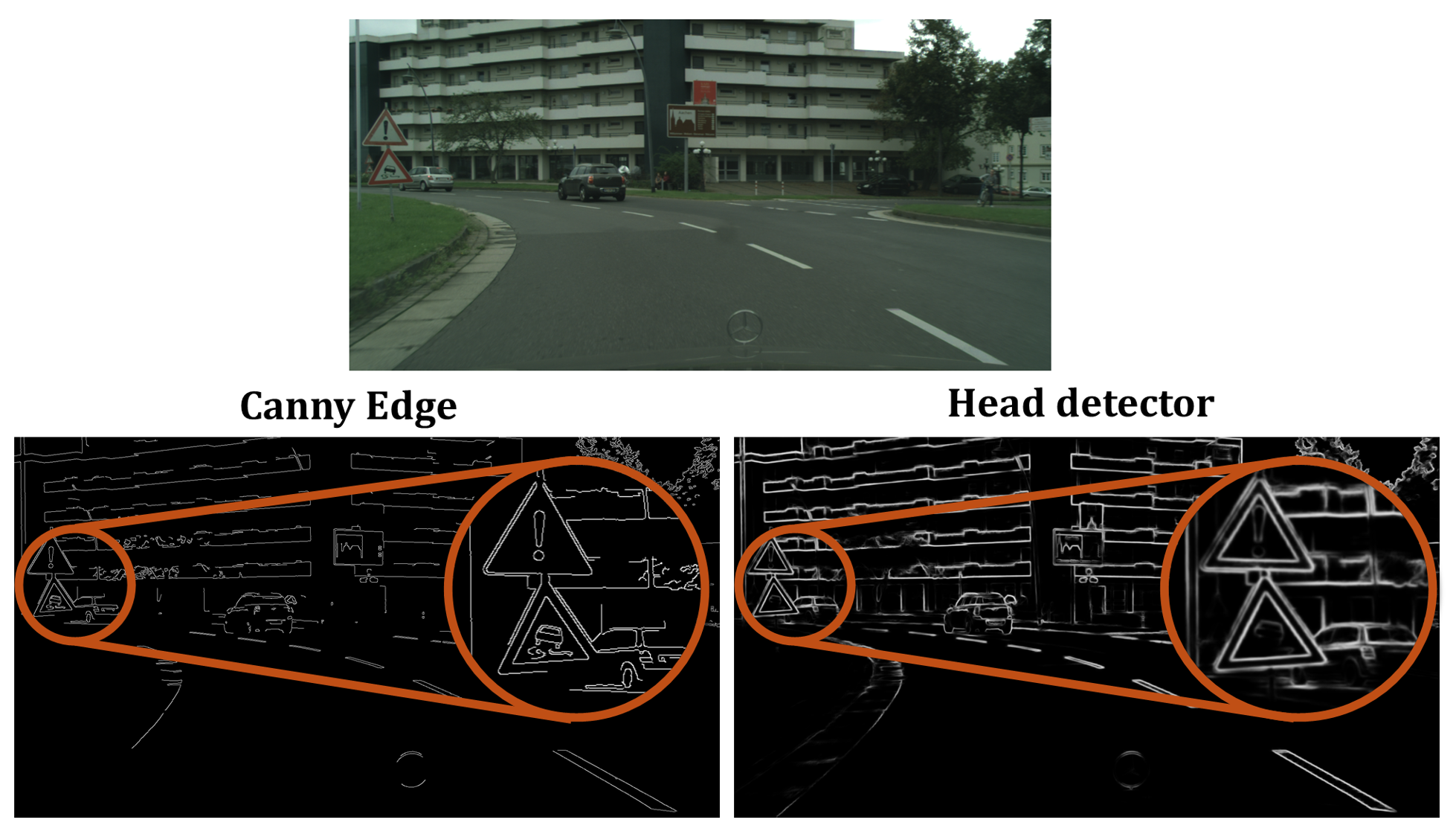}
\caption{Comparison of Canny and HED-style edge extraction. Canny preserves fine sign contours and thin structures more explicitly in this example, which is useful when the goal is structural preservation for driving scenes.}
\label{fig:canny_edge_compare}
\end{figure}

\textbf{Edges.} Edges represent intensity changes and fine contours. Canny is a classical deterministic algorithm \cite{canny1986edge}. HED and PiDiNet are learning-based edge detectors that emphasize perceptual contours \cite{xie2015hed,su2021pidinet}. In autonomous-driving images, however, thin structures such as distant signs, lane boundaries, poles, and vehicles are important. Canny often preserves these thin contours sharply. Therefore, this work uses Canny edges as the edge condition. \Cref{fig:canny_edge_compare} shows an example in which Canny keeps fine triangular sign contours more distinctly than a HED-style detector.

\textbf{Object and drivable-region evaluation.} Object preservation is evaluated through open-vocabulary detection with Grounding DINO \cite{liu2024groundingdino}. Classical and transformer-based detectors such as Faster R-CNN and DETR provide the background for this evaluation direction \cite{ren2015fasterrcnn,carion2020detr}. Drivable area is not evaluated by a separate model in the main tables because the Cityscapes-style semantic-label space already includes the road class; it is treated as part of semantic structure.

\subsection{Attention, tri-attention, and discrete selection}

Transformer attention computes relationships among Query, Key, and Value matrices \cite{vaswani2017attention}:
\begin{equation}
\mathrm{Attention}(Q,K,V)=\mathrm{softmax}\left(\frac{QK^\top}{\sqrt{d_k}}\right)V.
\end{equation}
This mechanism is widely used for visual-language models and conditional generation. CLIP maps images and text into a shared embedding space \cite{radford2021clip}, and Latent Diffusion Models use cross-attention to inject text conditions into generation \cite{rombach2022ldm}. FiLM provides another mechanism by modulating features channel-wise with condition-dependent scale and bias \cite{perez2018film}.

Standard attention computes a score from a two-way relation between query and key. Yu \etal{} point out that this bi-attention form cannot directly model three-way interactions and introduce Tri-Attention with an explicit context element \cite{yu2022triattention}. This work interprets the context as condition interaction among semantic segmentation, depth, and edge features. Rather than using tri-attention to continuously blend all conditions, \pamshort{} uses it to compute condition scores that decide which condition should dominate at each local feature position.

Discrete selection creates a differentiability problem. Suppose a local input $u_p$ produces scores $s_p=f_\theta(u_p)=[s_{p,1},\ldots,s_{p,K}]^\top$. Hard selection uses
\begin{equation}
 i_p^\star=\arg\max_k s_{p,k},\qquad g_{p,k}=\mathbf{1}[k=i_p^\star],\qquad \tilde z_p=\sum_{k=1}^K g_{p,k}z_{p,k}.
\end{equation}
Because $\arg\max$ is piecewise constant, $\partial g_{p,j}/\partial s_{p,k}=0$ almost everywhere, so the score generator receives no useful gradient. The Straight-Through Estimator (STE) addresses this by using hard values in the forward pass and soft surrogate gradients in the backward pass \cite{bengio2013stochastic,yin2019ste}. A common implementation is
\begin{equation}
 w = \mathrm{stopgrad}(w_\mathrm{hard}-w_\mathrm{soft})+w_\mathrm{soft}.
\end{equation}
In the forward pass, $w=w_\mathrm{hard}$. In the backward pass, the gradient flows through $w_\mathrm{soft}$. STE is widely used in quantized and conditional-computation settings, including recent low-bit language-model training \cite{wang2023bitnet,esser2025silq,malinovskii2024pvtuning}.

\subsection{Evaluation axes}

This work evaluates generated images along four axes. Structure preservation compares common projections of the original and generated image: semantic mIoU, depth RMSE, edge L1 error, and object-preservation F1. Realism uses CLIP-CMMD, a CLIP-embedding MMD metric motivated by limitations of FID in modern text-to-image evaluation \cite{heusel2017fid,binkowski2018kid,kynkaanniemi2019precisionrecall,gretton2012mmd,jayasumana2024cmmd}. Diversity uses AlexNet-LPIPS and $1-\mathrm{MS\text{-}SSIM}$ \cite{krizhevsky2012alexnet,zhang2018lpips,wang2003msssim,wang2004ssim}. Text alignment uses CLIP-based R-Precision, following the retrieval-based idea used in AttnGAN and related text-to-image evaluation \cite{radford2021clip,xu2018attngan,hessel2021clipscore}.

\section{Method}
\label{sec:method}

\subsection{Problem formulation}

Let $X$ be an original RGB driving image. Let $C^{\mathrm{seg}}$, $C^{\mathrm{dep}}$, and $C^{\mathrm{edge}}$ be the semantic segmentation map, depth map, and edge map extracted from $X$. Let $p$ be a text prompt that specifies target appearance changes such as weather and time of day. The objective is to learn or adapt a generator
\begin{equation}
 F(X)=G\left(X,C^{\mathrm{seg}},C^{\mathrm{dep}},C^{\mathrm{edge}},p\right)
\end{equation}
that changes appearance while preserving the original structural information needed by downstream annotations.

\subsection{Overall pipeline}

\begin{figure}[tbp]
\centering
\includegraphics[width=0.95\linewidth]{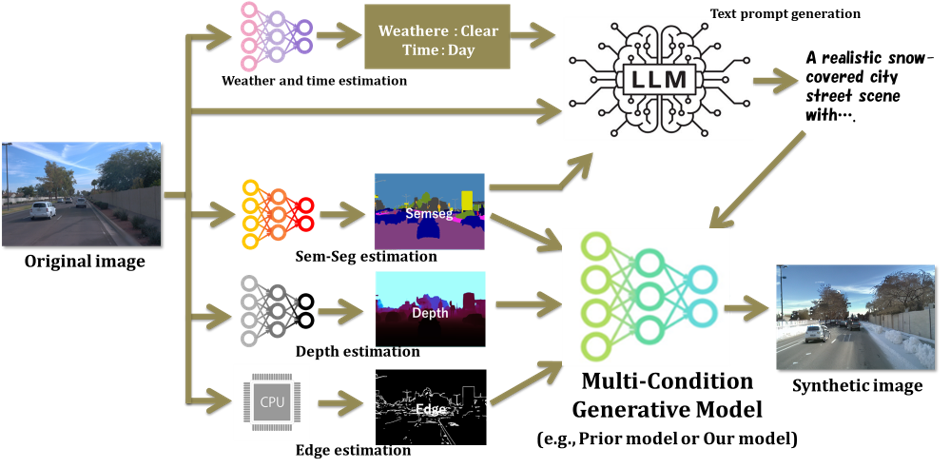}
\caption{Overall multi-condition generation pipeline. The input RGB image is converted into weather/time estimates, semantic segmentation, depth, and edges. A prompt is generated to change appearance, and the prompt plus local conditions are passed to the multi-condition generator.}
\label{fig:pipeline}
\end{figure}

\Cref{fig:pipeline} shows the proposed pipeline. From the input RGB image, the system estimates weather and time using CLIP-like image-text similarity. It then uses an LLM/VLM-based prompt process to generate a text prompt that can explicitly change weather, time, or both. In parallel, OneFormer estimates semantic segmentation, Metric3Dv2 estimates depth, and Canny computes edges. These three local conditions and the generated prompt are fed into the multi-condition generation model. The output is a synthetic image that preserves the original road-scene structure while changing appearance.

\subsection{Prompt generation}
\label{sec:prompt_generation}

\begin{figure}[tbp]
\centering
\includegraphics[width=0.95\linewidth]{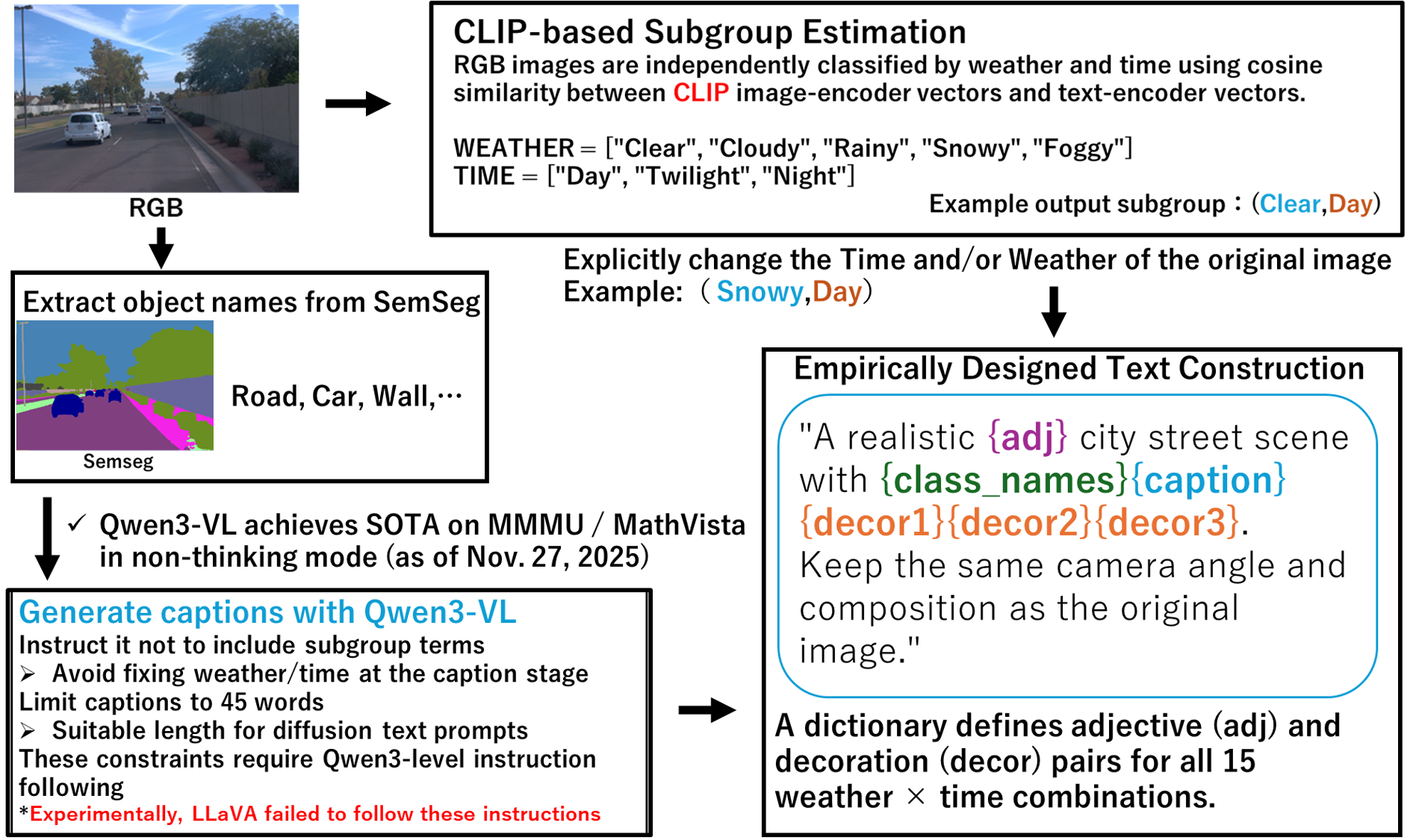}
\caption{Prompt-generation pipeline. CLIP estimates the source weather and time subgroup. Semantic segmentation supplies object names. Qwen3-VL generates a caption without explicit weather/time words. A style dictionary then adds target weather/time adjectives and decorations to build the final prompt.}
\label{fig:text_prompt_pipeline}
\end{figure}

The inference-time prompt process separates a layout caption from a target style description. This separation is important because the caption should describe objects, spatial relations, background, viewpoint, and image quality without fixing weather or time-of-day words, while the target style should specify the intended weather and time. \Cref{fig:text_prompt_pipeline} summarizes the process.

First, a VLM, Qwen3-VL-32B-Instruct, receives the image and the object classes extracted from semantic segmentation. It produces a concise one- or two-sentence caption, limited to roughly 45 words, and is instructed not to mention rain, snow, fog, clear/sunny, day, night, dawn, twilight, or similar terms. This keeps layout information separate from target appearance.

Second, CLIP estimates the source subgroup $(w_{\mathrm{src}},t_{\mathrm{src}})$ from the weather labels $\{\mathrm{Clear},\mathrm{Cloudy},\mathrm{Rainy},\mathrm{Snowy},\mathrm{Foggy}\}$ and the time labels $\{\mathrm{Day},\mathrm{Twilight},\mathrm{Night}\}$. At inference, the target subgroup is sampled from labels excluding the source label on each axis:
\begin{equation}
 w_{\mathrm{tgt}}\sim\mathrm{Uniform}(W\setminus\{w_{\mathrm{src}}\}),\qquad
 t_{\mathrm{tgt}}\sim\mathrm{Uniform}(T\setminus\{t_{\mathrm{src}}\}).
\end{equation}
This setting changes both weather and time.

Third, the target subgroup indexes a style dictionary containing one adjective and three decoration phrases for each of the $5\times 3=15$ weather-time combinations. The final prompt template is
\begin{quote}\small
\texttt{A realistic \{adj\} city street scene with \{class\_names\}. \{caption\} \{decor1\} \{decor2\} \{decor3\}. Keep the same camera angle and composition as the original image.}
\end{quote}
For example, if the target subgroup is Rainy-Day, the adjective can be \texttt{rain-soaked}, and the decorations can describe wet asphalt, raindrops on windows, and softened edges due to light drizzle. The caption preserves layout; the dictionary changes appearance.

\subsection{Base generator}
\label{sec:base_generator}

The base generator $G_{\mathrm{base}}$ uses Stable Diffusion 1.5 style latent diffusion, text conditioning, and Uni-ControlNet-compatible multi-stage local injection. The core trainable component is a local control branch initialized from Uni-ControlNet. Although the actual conditions are edge, depth, and semantic segmentation, implementation compatibility is maintained by embedding them into the seven-condition, 21-channel Uni-ControlNet format:
\begin{equation}
 C^{\mathrm{uni}} = \left(C^{\mathrm{edge}},0,0,0,0,C^{\mathrm{dep}},C^{\mathrm{seg}}\right).
\end{equation}
Unused condition slots are filled with zeros. A local encoder $E_{\mathrm{loc}}$ maps this tensor into four local feature levels:
\begin{equation}
 \{H^{(m)}\}_{m=1}^4 = E_{\mathrm{loc}}(C^{\mathrm{uni}}).
\end{equation}
A local control backend $B_{\mathrm{loc}}$ converts these features into control tensors matched to skip and middle structures of the fixed diffusion U-Net:
\begin{equation}
 \{\Delta^{(r)}\}_{r=1}^R = B_{\mathrm{loc}}(z_t,t,c,\{H^{(m)}\}_{m=1}^4),
\end{equation}
where $z_t$ is the noisy latent at timestep $t$ and $c$ is the text context. The final noise prediction is
\begin{equation}
 \hat\epsilon_\theta = U_{\mathrm{SD}}(z_t,t,c,\{\Delta^{(r)}\}_{r=1}^R).
\end{equation}
During training, the Stable Diffusion denoising U-Net, VAE, and text encoder are frozen. The local control branch is updated, so the model adapts local-condition interpretation without destroying the pretrained text-to-image prior. The global control branch is kept for compatibility, but its input is fixed to a zero vector.

\begin{figure}[tbp]
\centering
\includegraphics[width=0.92\linewidth]{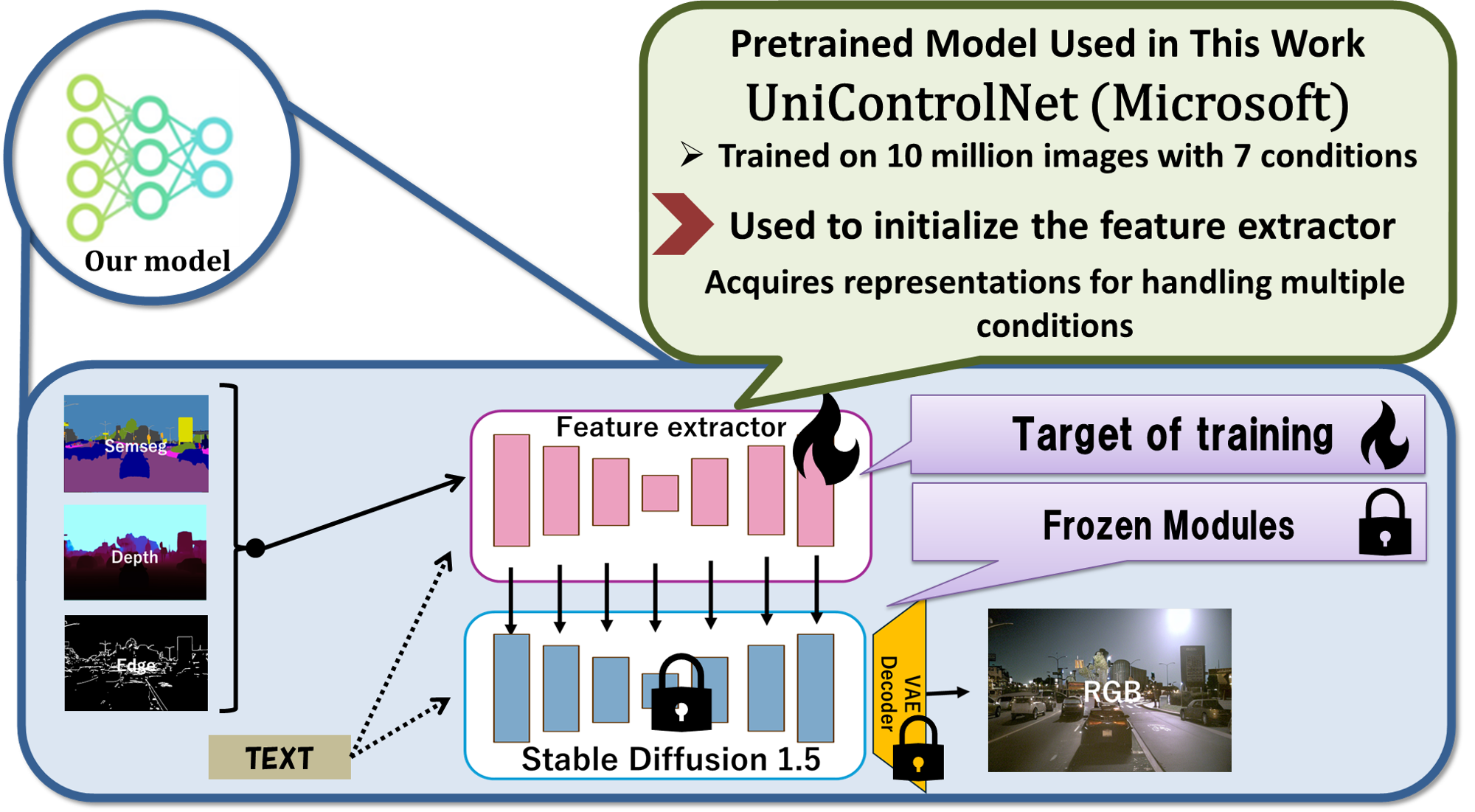}
\caption{Model detail. The feature extractor is initialized from Uni-ControlNet local-control weights and processes three structural conditions. Stable Diffusion, the VAE, and the text encoder are frozen, so training mainly adapts the local control branch.}
\label{fig:model_detail}
\end{figure}

\subsection{Patch-wise Adaptation Module}
\label{sec:pam}

The base generator feeds all three conditions into a shared local feature extractor. This is compatible with Uni-ControlNet but mixes depth, edges, and segmentation early in the feature space. Depth imposes low-frequency geometry, edges impose high-frequency contours, and semantic segmentation imposes region consistency. When these signals disagree locally, roads may lose continuity, object shapes may collapse, or distant structures may hallucinate. \pamshort{} is introduced to reduce this conflict.

\pamshort{} replaces only the early shared local extractor with condition-specific stems and a local selection mechanism. The later control backend and the fixed Stable Diffusion backbone remain the same as in the base generator. Let $S_{\mathrm{edge}}$, $S_{\mathrm{dep}}$, and $S_{\mathrm{seg}}$ be condition-specific stems. They map the three conditions to a $1/8$-resolution local feature grid:
\begin{equation}
 F^k=S_k(C^k),\qquad k\in\{\mathrm{edge},\mathrm{dep},\mathrm{seg}\}.
\end{equation}
Here, ``patch-wise'' means each position in this local feature grid, not an explicit non-overlapping image patch in pixel space.

At spatial location $u$, form a condition-token sequence
\begin{equation}
 Z_u=\left[f_u^{\mathrm{edge}},f_u^{\mathrm{dep}},f_u^{\mathrm{seg}}\right].
\end{equation}
A context vector is computed from the timestep embedding and a pooled text context:
\begin{equation}
 q=\psi_t(t)+\psi_c(\mathrm{Pool}(c)).
\end{equation}
A tri-attention-style gate computes condition-interaction features and scores:
\begin{equation}
 \hat Z_u=T(Z_u,q),\qquad s_u=g(\hat Z_u).
\end{equation}
The tri-attention block is not used to continuously fuse all features. It is used as a score generator to decide which condition should be selected.

\begin{figure}[tbp]
\centering
\includegraphics[width=0.92\linewidth]{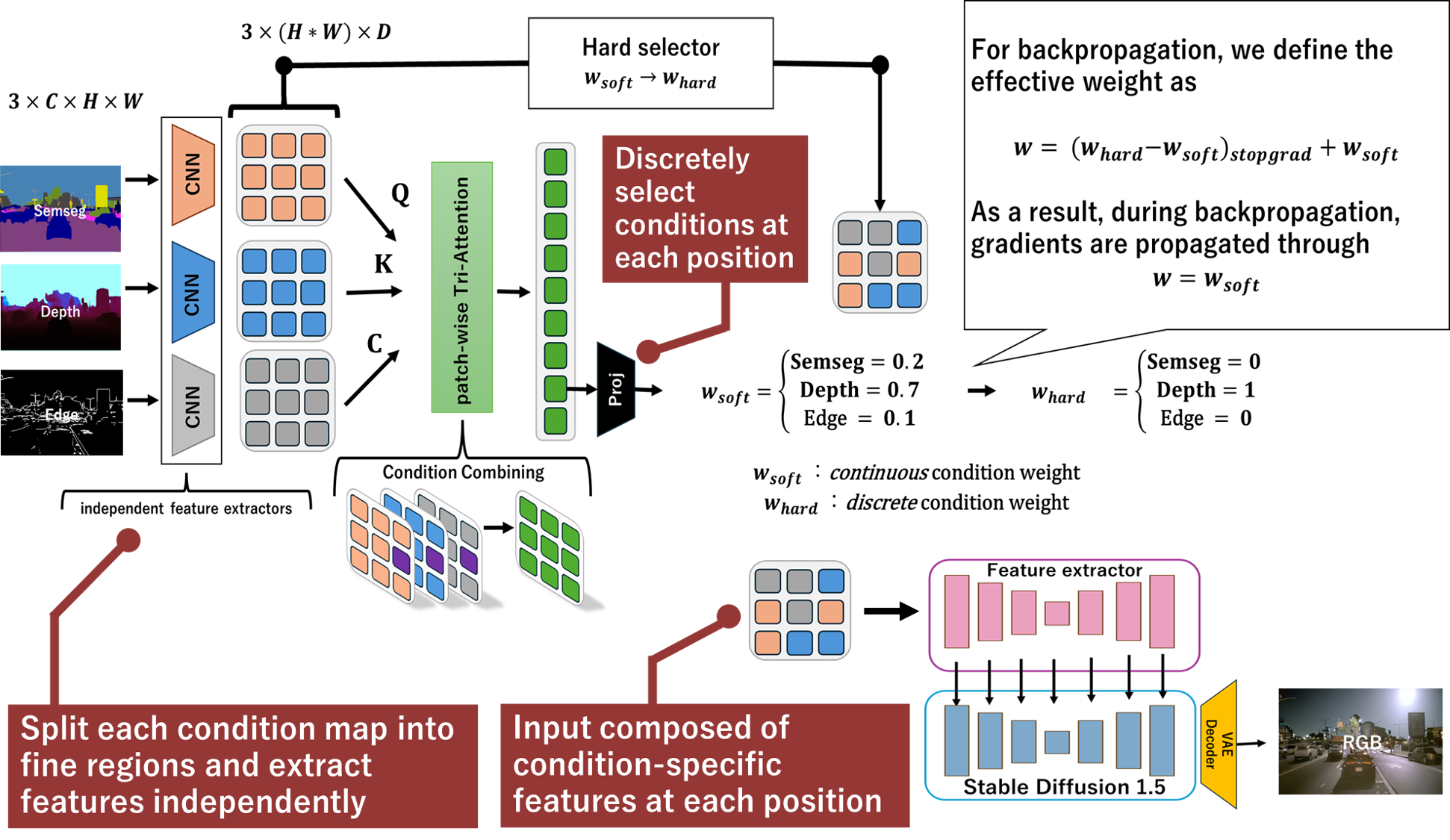}
\caption{Patch-wise Adaptation Module. Edge, depth, and semantic segmentation are processed by condition-specific stems. At each local feature-grid position, a tri-attention-style gate scores the three conditions, and a straight-through hard selector chooses the locally effective condition before multi-stage control injection.}
\label{fig:pam_architecture}
\end{figure}

For hard trainable selection, define
\begin{equation}
 \pi_u=\mathrm{softmax}(s_u/\tau),\qquad y_u=\mathrm{onehot}\left(\arg\max_k s_{u,k}\right),
\end{equation}
then use the straight-through weight
\begin{equation}
 w_u=\mathrm{stopgrad}(y_u-\pi_u)+\pi_u.
\end{equation}
The selected local feature is
\begin{equation}
 \tilde f_u=\sum_{k\in\{\mathrm{edge},\mathrm{dep},\mathrm{seg}\}} w_{u,k}f_u^k.
\end{equation}
The selected features are reassembled into a grid $\tilde F$, optionally enhanced by a lightweight residual block, and then passed to the same tail extractor and local control backend as the base generator. \Cref{fig:pam_architecture} illustrates the design.

\section{Experiments}
\label{sec:experiments}

\subsection{Training and evaluation data}

Training uses a combined autonomous-driving-related dataset shown in \Cref{tab:training_data}. It includes BDD10K segmentation, Cityscapes, GTA5 Synthetic Dataset, nuImages front camera images, and BDD100K images excluding the BDD10K overlap. The total training set contains 146,809 images. Evaluation uses 3,048 Waymo Open Dataset front-camera images. Three frames are extracted from each segment: the first frame, a frame around the middle of the segment, and the last frame. This reduces redundancy while retaining temporal variation across each scene.

\begin{table}[tbp]
\centering
\caption{Training datasets.}
\label{tab:training_data}
\begin{tabular}{lrr}
\toprule
Dataset & Images & Reference \\
\midrule
BDD10K Segmentation & 8,000 & \cite{yu2020bdd100k} \\
Cityscapes & 3,475 & \cite{cordts2016cityscapes} \\
GTA5 Synthetic Dataset & 24,966 & \cite{richter2016playingfordata} \\
nuImages (front camera) & 18,368 & \cite{caesar2020nuscenes} \\
BDD100K (excluding BDD10K overlap) & 92,000 & \cite{yu2020bdd100k} \\
\midrule
Total & 146,809 & \\
\bottomrule
\end{tabular}
\end{table}

\begin{table}[tbp]
\centering
\caption{Evaluation dataset.}
\label{tab:evaluation_data}
\begin{tabular}{lrr}
\toprule
Dataset & Images & Reference \\
\midrule
Waymo Open Dataset (front camera) & 3,048 & \cite{sun2020waymo} \\
\bottomrule
\end{tabular}
\end{table}

Inputs are resized to $512\times512$. During inference, DDIM sampling uses 50 steps. The main training and inference hyperparameters are shown in \Cref{tab:hyperparams}. Experiments compare 30K, 60K, and 90K training steps. The batch size is 4. The optimizer is AdamW with learning rate $10^{-5}$. Stable Diffusion is frozen; the local adapter and \pamshort{} parameters are trained.

\begin{table}[tbp]
\centering
\caption{Main training and inference hyperparameters.}
\label{tab:hyperparams}
\begin{tabular}{ll}
\toprule
Item & Setting \\
\midrule
Optimizer & AdamW \\
Learning rate & $1\times10^{-5}$ \\
Batch size & 4 \\
Training steps & 30K / 60K / 90K \\
Gradient checkpointing & Enabled \\
Stable Diffusion update & Disabled (frozen) \\
Training target & Local adapter + \pamshort{} \\
Inference resolution & 512 \\
DDIM steps & 50 \\
CFG scale & 7.5 \\
Local-condition strength & 1.0 \\
\bottomrule
\end{tabular}
\end{table}

Compared with prior multi-condition generation research, the training set is relatively small. Uni-ControlNet samples 10 million image-text pairs from LAION for training, and PixelPonder uses 2.5 million samples from MultiGen-20M \cite{zhao2023unicontrolnet,pan2025pixelponder}. By contrast, this work uses about 150K training images. The evaluation set of 3,048 Waymo images is close to the scale of common 5K-image validation sets and is sufficient for quantitative comparison. The experiments therefore test whether structure-preserving multi-condition generation can be effective under a more limited but domain-relevant training scale.

\subsection{Model variants}

\Cref{tab:model_names} defines the model names used in the experiments. One training step corresponds to one batch of four images. Therefore, 30K, 60K, and 90K steps correspond to roughly 120K, 240K, and 360K image presentations.

\begin{table}[tbp]
\centering
\caption{Model variants.}
\label{tab:model_names}
\resizebox{\textwidth}{!}{%
\begin{tabular}{p{35mm}p{120mm}}
\toprule
Model & Description \\
\midrule
FullScratch30K & Stable Diffusion plus the Uni-ControlNet-style local branch trained for 30K steps from random initialization, without Uni-ControlNet pretrained local-control weights. \\
Tune30K & Model initialized with public Uni-ControlNet-compatible weights and trained for 30K steps. \\
Tune60K & Same as Tune30K, trained for 60K steps. \\
Tune90K & Same as Tune30K, trained for 90K steps. \\
PAM60K & Uni-ControlNet-initialized model with \pamshort{} added, trained for 60K steps. \\
Uni-ControlNet & Public Uni-ControlNet model. It receives the same edge, depth, and segmentation conditions but is not fine-tuned for the autonomous-driving domain. \\
DGInStyle & Public DGInStyle model. It receives semantic segmentation only, because DGInStyle is semantic-mask conditioned. The same text prompt is used for fair appearance control. \\
\bottomrule
\end{tabular}}
\end{table}

\subsection{Evaluation metrics}
\label{sec:metrics}

The primary goal is structural consistency with the original image. However, if only structure is optimized, images can become unrealistic, overly similar, or weakly aligned with the text prompt. Therefore, the evaluation uses four axes.

\begin{figure}[tbp]
\centering
\includegraphics[width=0.92\linewidth]{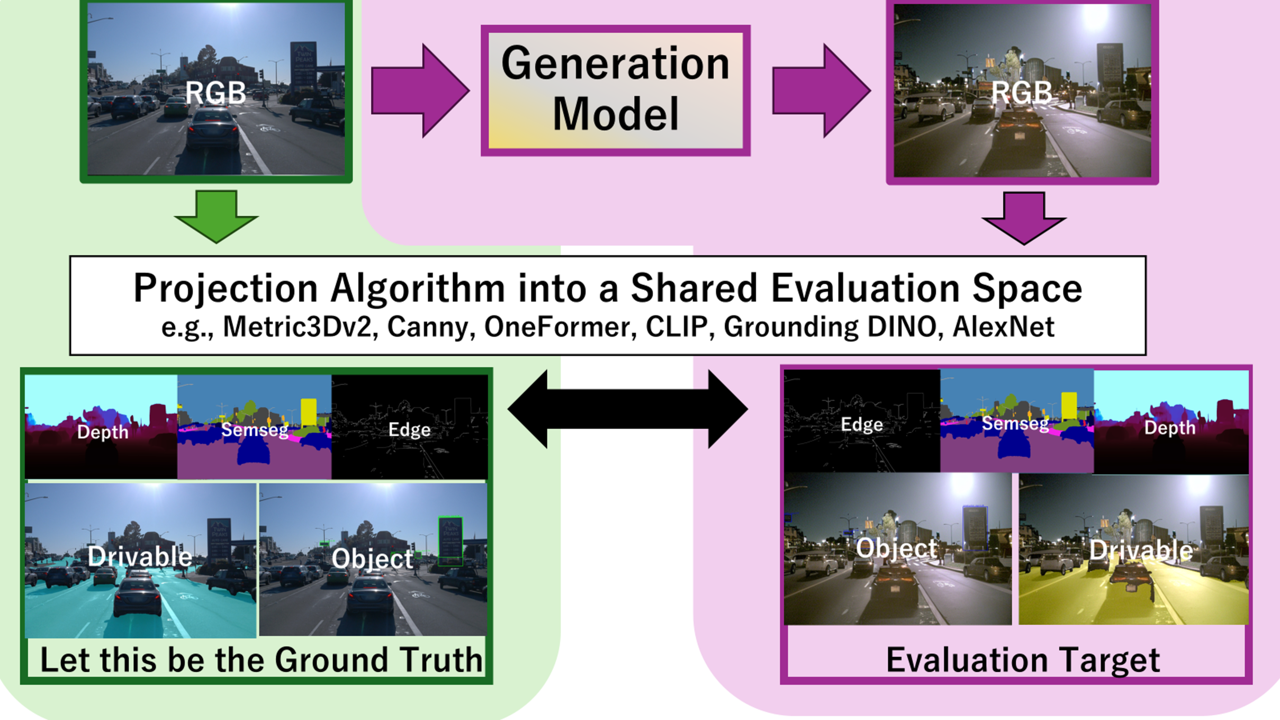}
\caption{Common-projection structure evaluation. The original image and generated image are passed through the same structural projectors. Original-side outputs are treated as pseudo ground truth, and generated-side outputs are compared in semantic, depth, edge, and object spaces rather than in raw RGB pixel space.}
\label{fig:evaluation_structure}
\end{figure}

\textbf{Structure preservation.} Original and generated images are projected into common evaluation spaces using OneFormer, Metric3Dv2, Canny, and Grounding DINO. The original-side prediction is treated as pseudo ground truth, and the generated-side prediction is compared with it. The main metrics are semantic segmentation mIoU, depth RMSE, masked edge L1 error, and traffic-object preservation F1. Larger is better for mIoU and F1; smaller is better for RMSE and L1. \Cref{fig:evaluation_structure} shows the intuition.

\begin{figure}[tbp]
\centering
\includegraphics[width=0.92\linewidth]{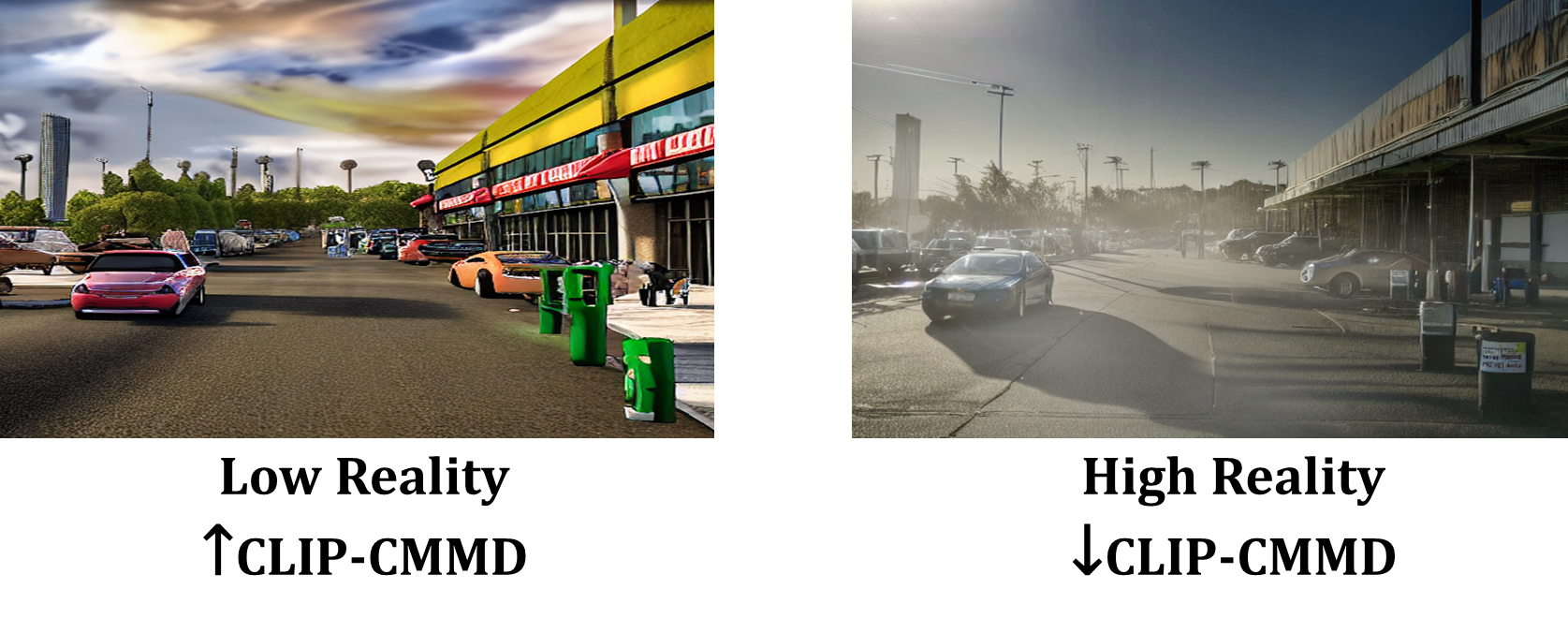}
\caption{Intuition for realism evaluation. Low-realism images may preserve layout but show unnatural texture, color, or lighting. Higher-realism images show more plausible light, shadow, and atmosphere. CLIP-CMMD is smaller for the latter.}
\label{fig:reality_example}
\end{figure}

\textbf{Realism.} CLIP-CMMD measures the MMD distance between original and generated image distributions in CLIP image-embedding space \cite{radford2021clip,gretton2012mmd,jayasumana2024cmmd}. Smaller values indicate generated images closer to the real image distribution. \Cref{fig:reality_example} illustrates the intuition.

\begin{figure}[tbp]
\centering
\includegraphics[width=0.92\linewidth]{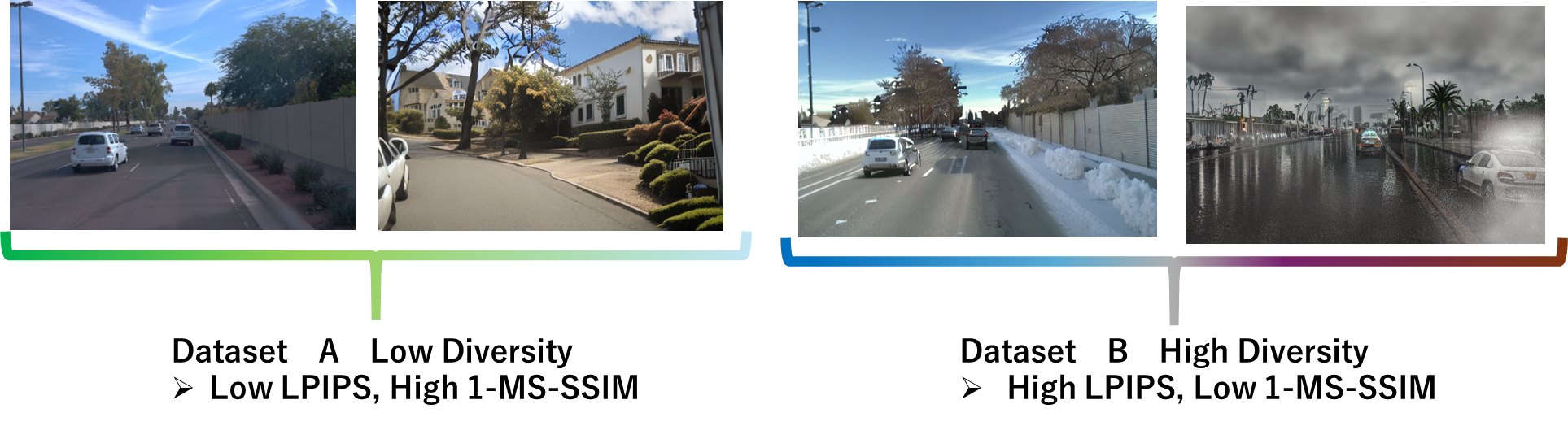}
\caption{Intuition for diversity evaluation. Similar generated image pairs have small distances, while pairs with substantially different weather, road condition, and illumination have larger distances. LPIPS and $1-\mathrm{MS\text{-}SSIM}$ quantify this separation.}
\label{fig:diversity_example}
\end{figure}

\textbf{Diversity.} The diversity of generated images is evaluated with AlexNet-LPIPS and $1-\mathrm{MS\text{-}SSIM}$ on randomly sampled image pairs. Larger values indicate more diverse generated appearances. \Cref{fig:diversity_example} shows the intuition.

\begin{figure}[tbp]
\centering
\includegraphics[width=0.92\linewidth]{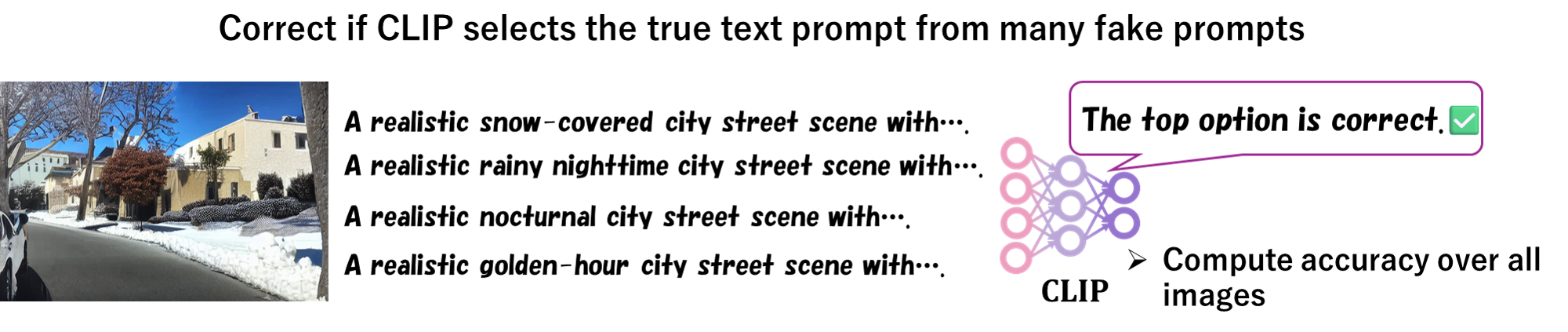}
\caption{Intuition for text-alignment evaluation. The correct prompt and 99 mismatched prompts are ranked by CLIP similarity to the generated image. R-Precision is high when the correct prompt appears near the top.}
\label{fig:text_alignment_example}
\end{figure}

\textbf{Text alignment.} CLIP-R-Precision evaluates whether the correct prompt is ranked among 99 mismatched prompts. A generated image and 100 candidate prompts are embedded with CLIP, ranked by similarity, and a hit is counted if the matched prompt appears in the top $K$. Larger values indicate stronger text alignment. \Cref{fig:text_alignment_example} shows the intuition.

\section{Results}
\label{sec:results}

\subsection{Effect of pretraining}

\Cref{tab:pretraining_effect} compares FullScratch30K and Tune30K. Except for edge L1 and a small realism difference, Uni-ControlNet initialization improves the metrics. Semantic mIoU improves from 0.2445 to 0.2776. Depth RMSE improves from 40.41 to 36.72. Object-preservation F1 increases from 0.0455 to 0.1369, which is the largest relative difference. Diversity and text alignment also improve. This indicates that pretrained local-control representations help the model interpret multiple structural conditions and inject them into the diffusion U-Net. Object preservation benefits especially because preserving a traffic object requires not only a class region but also a plausible local shape and depth/edge consistency.

\begin{table}[tbp]
\centering
\caption{Effect of Uni-ControlNet initialization at 30K steps. Relative change is computed with the pretrained model as the reference, matching the thesis convention. For upward metrics, negative change means the no-pretraining model is worse; for downward metrics, positive change means the no-pretraining model is worse.}
\label{tab:pretraining_effect}
\resizebox{\textwidth}{!}{%
\begin{tabular}{llllrrl}
\toprule
Category & Metric & Direction & With pretraining & Without pretraining & $\Delta$ [\%] & Note \\
\midrule
Semantic Segmentation & mIoU & $\uparrow$ & 0.2776 & 0.2445 & -11.9 & Pretraining better \\
Depth & RMSE & $\downarrow$ & 36.72 & 40.41 & +10.0 & Pretraining better \\
Edge & L1 Error & $\downarrow$ & 0.05038 & 0.03759 & -25.4 & Scratch better on edge \\
Object Preservation & F1 Score & $\uparrow$ & 0.1369 & 0.0455 & -66.8 & Pretraining much better \\
Diversity & $1-\mathrm{MS\text{-}SSIM}$ & $\uparrow$ & 0.8740 & 0.8450 & -3.32 & Pretraining better \\
Reality & CLIP-CMMD & $\downarrow$ & 0.1881 & 0.1827 & -2.88 & Scratch slightly better \\
Text Alignment & R-Precision@1 & $\uparrow$ & 0.3468 & 0.2894 & -16.6 & Pretraining better \\
\bottomrule
\end{tabular}}
\end{table}

\begin{figure}[tbp]
\centering
\includegraphics[width=0.95\linewidth]{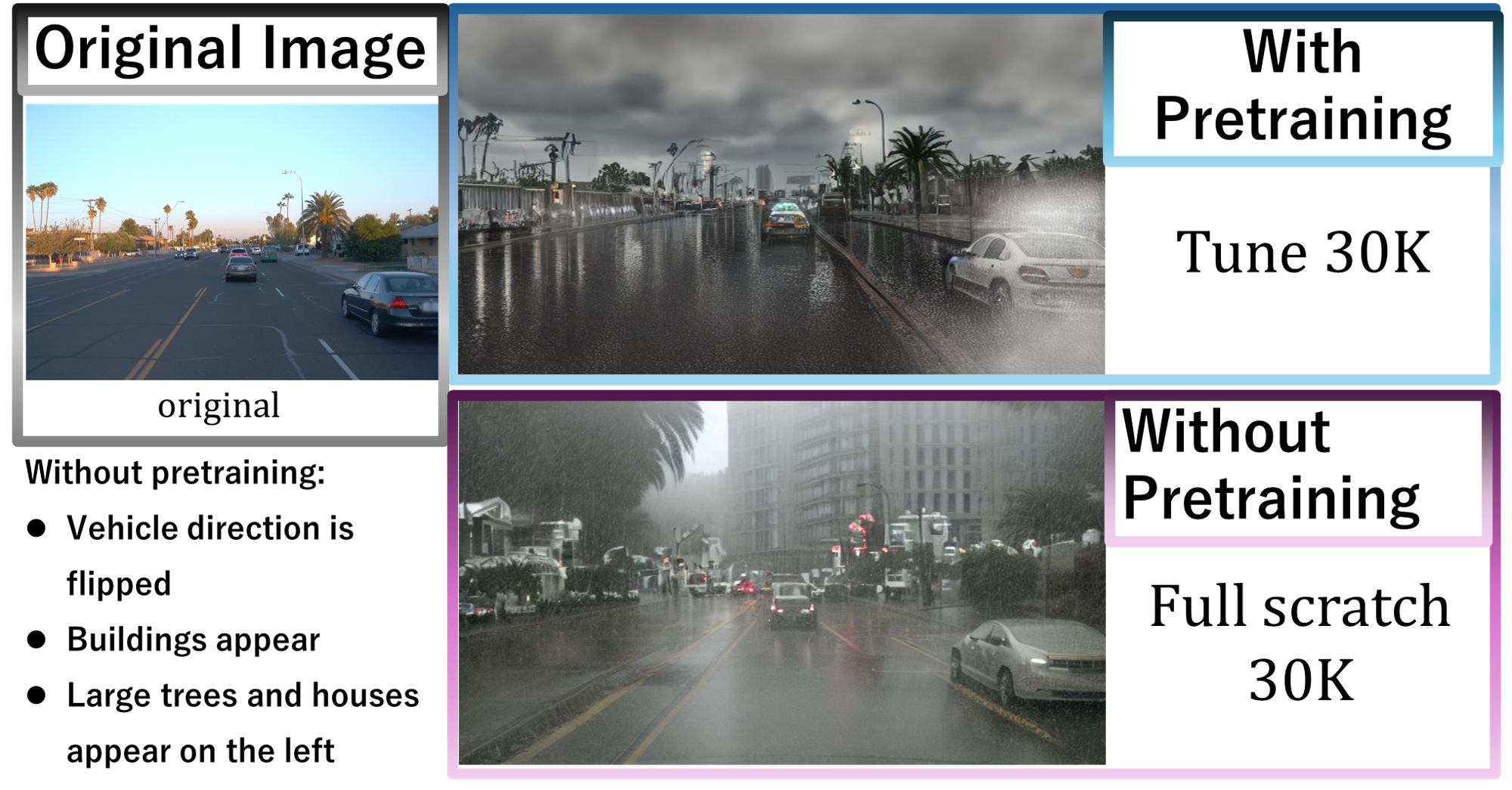}
\caption{Qualitative comparison of training with and without Uni-ControlNet initialization. Without pretraining, road structure and vehicle geometry can collapse or hallucinate. With pretraining, the generated image better preserves road shape and object placement.}
\label{fig:pretrain_impact}
\end{figure}

\Cref{fig:pretrain_impact} gives a qualitative example. Both scratch and pretrained models change the input from a twilight-like clear scene toward rainy daytime appearance. However, the scratch model flips the direction of the foreground car, introduces buildings and trees that are not present in the original image, and hallucinates distant traffic lights. The pretrained model preserves road shape and object placement more consistently. Therefore, pretraining improves the structural prior while keeping appearance-change ability.

The edge result is an exception. The scratch model has lower Canny L1 error, but this does not mean it preserves scene structure better overall. Edge L1 can be improved by producing sharper local contours even when global scene identity is wrong. The mismatch between this edge score and mIoU/depth/object scores illustrates why multiple structural metrics are necessary.

\subsection{Scaling with training steps}

\begin{figure}[tbp]
\centering
\includegraphics[width=0.95\linewidth]{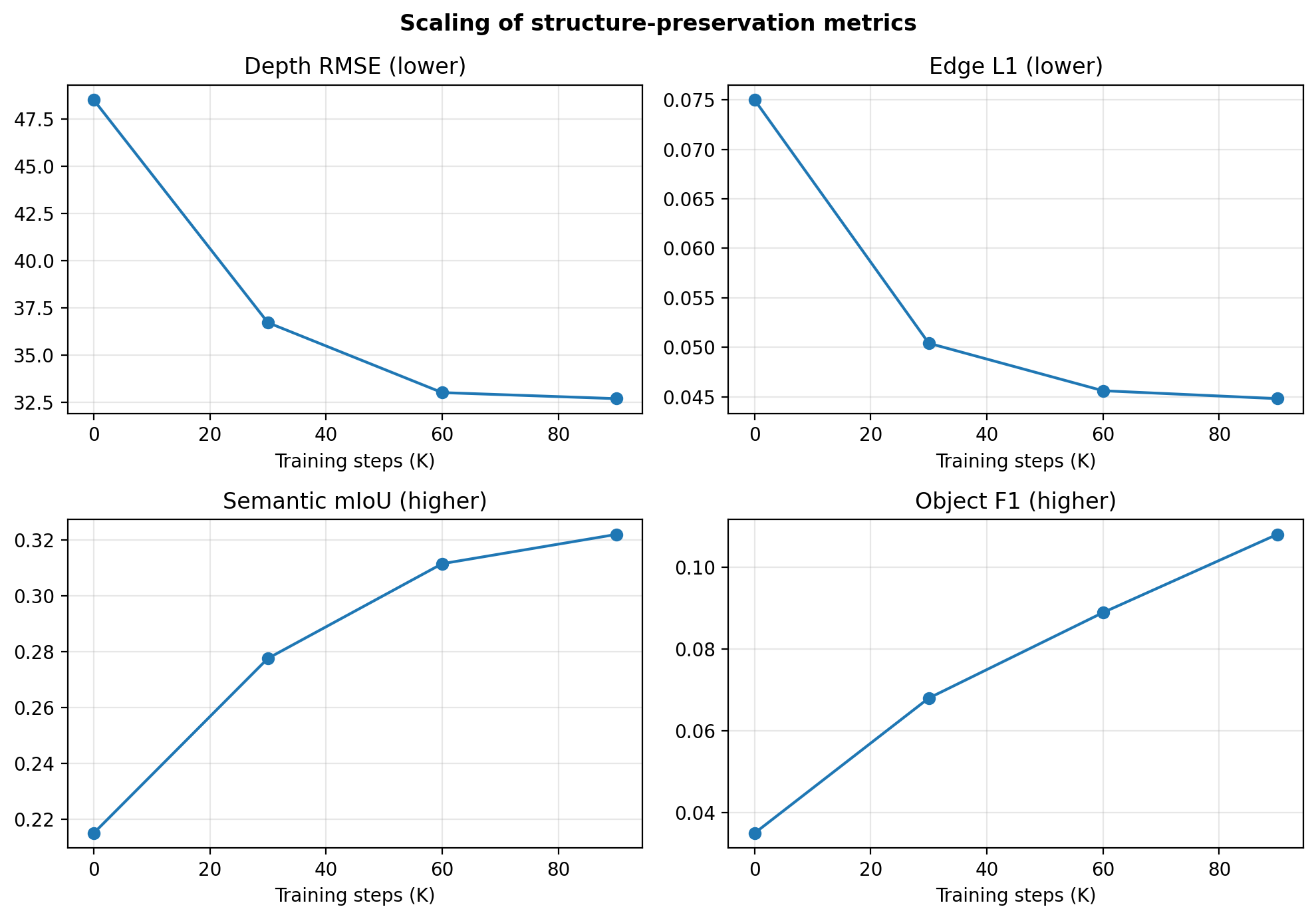}
\caption{Scaling behavior of structure-related metrics for Tune models. Depth RMSE and edge L1 are lower-is-better; semantic mIoU and object F1 are higher-is-better. Most improvements are largest between 0 and 30K, become milder from 30K to 60K, and approach saturation near 90K.}
\label{fig:scaling_structure}
\end{figure}

\begin{figure}[tbp]
\centering
\includegraphics[width=0.95\linewidth]{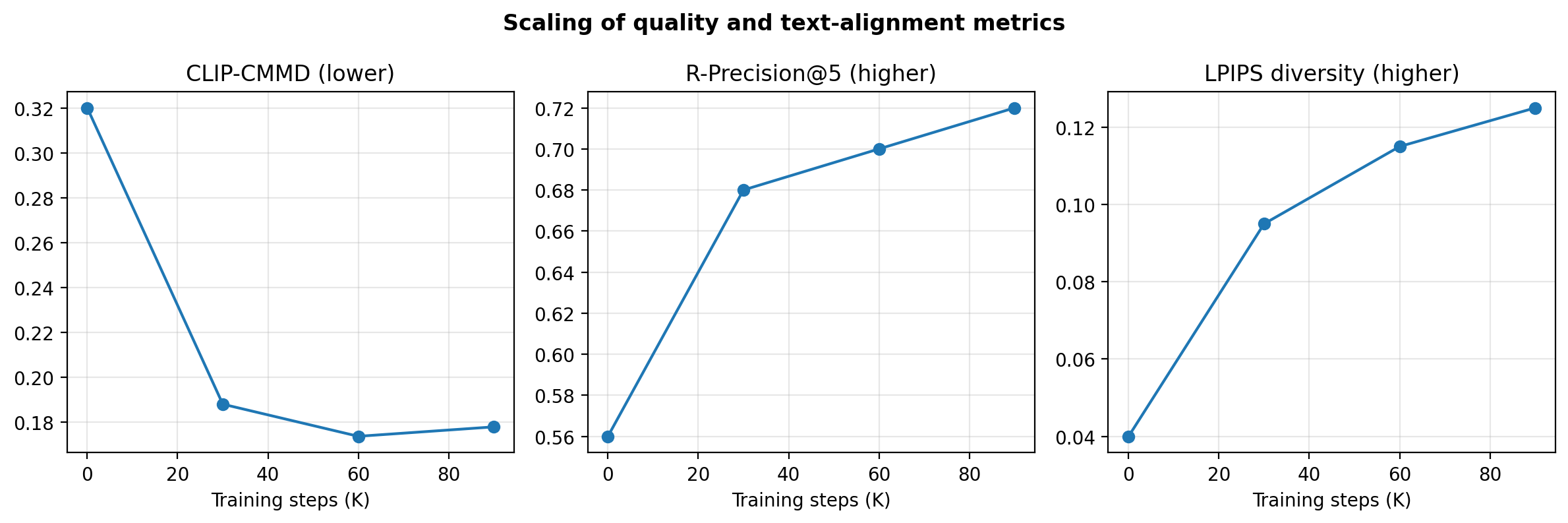}
\caption{Scaling behavior of quality, diversity, and text-alignment metrics for Tune models. CLIP-CMMD is lower-is-better, while R-Precision@5 and LPIPS are higher-is-better. Realism improves until 60K and slightly reverses at 90K.}
\label{fig:scaling_quality}
\end{figure}

\Cref{fig:scaling_structure} shows scaling of the Tune models. Semantic mIoU and object F1 increase monotonically from 0 to 30K to 60K to 90K, while depth RMSE and edge L1 decrease monotonically. Thus, structure preservation improves consistently as fine-tuning proceeds. \Cref{fig:scaling_quality} shows quality-related metrics. R-Precision@5 and LPIPS increase monotonically. CLIP-CMMD decreases until 60K and then slightly worsens at 90K. This suggests that 60K reaches a good balance for realism, while 90K continues improving some structural and diversity/text metrics but may begin to slightly restrict appearance freedom.

\begin{figure}[tbp]
\centering
\includegraphics[width=0.95\linewidth]{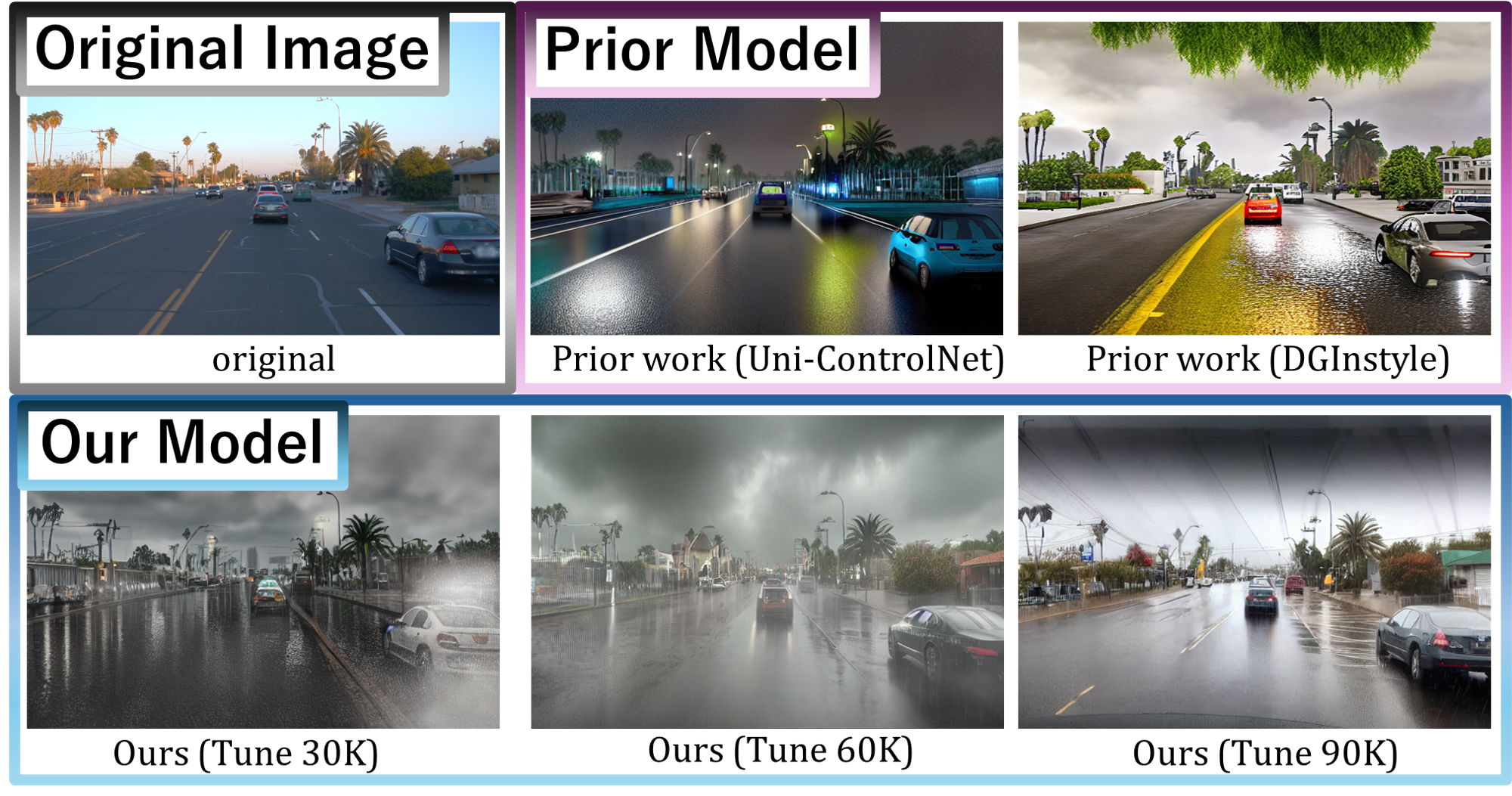}
\caption{Qualitative comparison across training steps. Prior-work baselines can show unrealistic style or structural changes, while the Tune models better preserve the original road structure and realism as training proceeds.}
\label{fig:qualitative_tune}
\end{figure}

\begin{figure}[tbp]
\centering
\includegraphics[width=0.95\linewidth]{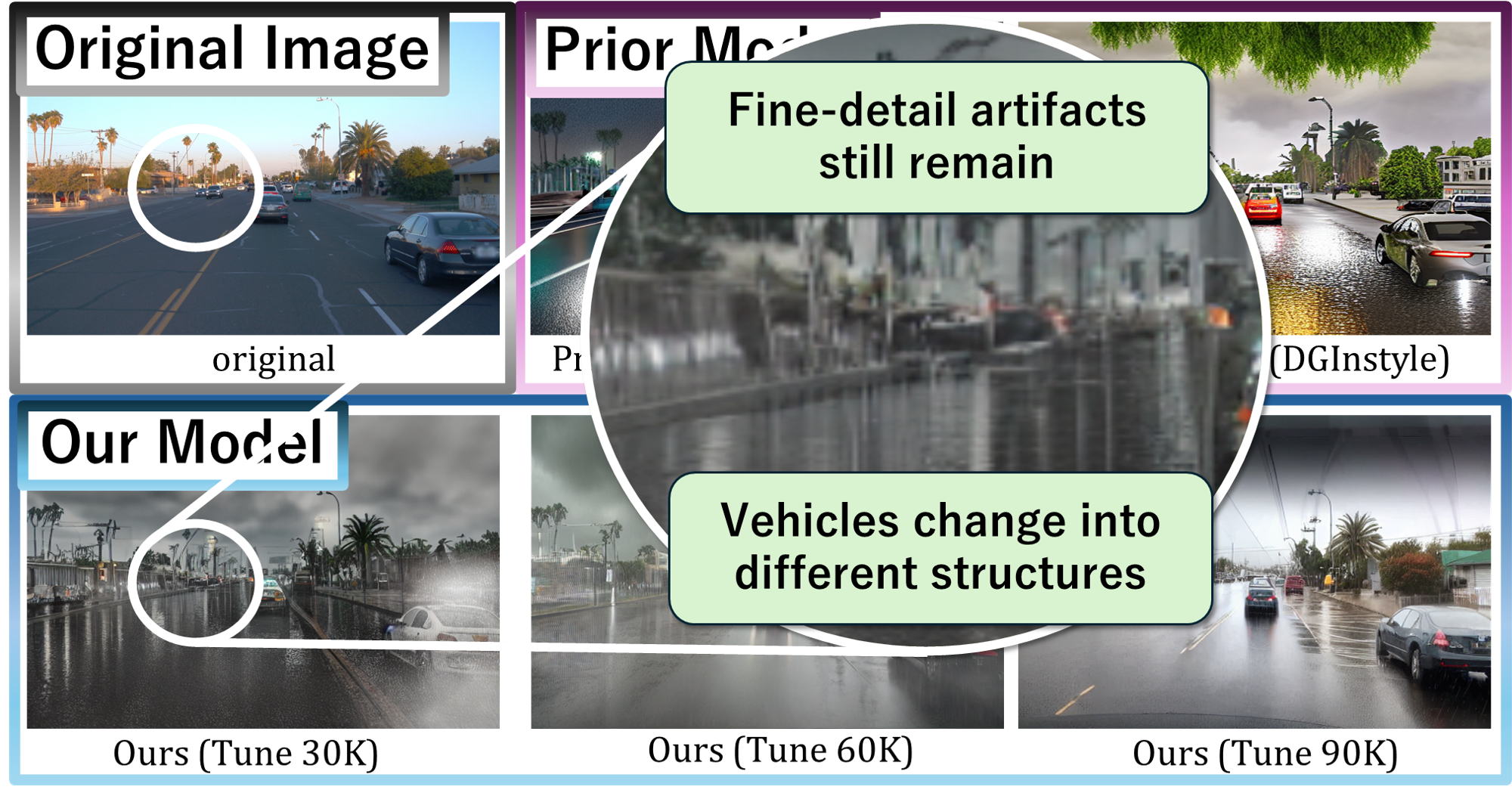}
\caption{Zoomed local artifact that can remain even when training steps are increased. A vehicle in the original can disappear or change into another structure, motivating local condition-conflict suppression.}
\label{fig:qualitative_tune_zoom}
\end{figure}

\Cref{fig:qualitative_tune} compares qualitative results across training progress and prior-work baselines. Uni-ControlNet can produce cyber-like reflections and distort the leading vehicle. DGInStyle can produce anime-like color and add vegetation not present in the original. The Tune models are more realistic and increasingly preserve road and vehicle layout as the number of steps grows. Nevertheless, local artifacts remain even after additional training. \Cref{fig:qualitative_tune_zoom} shows a case in which a vehicle in the original image disappears or changes into a different structure. This motivates \pamshort{}: even when global metrics improve, local condition conflict can still destroy annotation-relevant details.

\subsection{Effect of \pamshort{}}

\Cref{tab:pam_effect} compares Tune60K and PAM60K. \pamshort{} improves all four structural metrics. Semantic mIoU increases from 0.3115 to 0.3310. Depth RMSE drops from 33.02 to 27.77. Edge L1 decreases from 0.04561 to 0.04493. Object-preservation F1 increases from 0.0889 to 0.1071. The largest gains are depth and object preservation, which supports the hypothesis that local conflict among depth, edge, and segmentation is especially important for geometry and object existence.

\begin{table}[tbp]
\centering
\caption{Effect of \pamshort{} at 60K steps. Relative change is computed with PAM as the reference in the same thesis convention.}
\label{tab:pam_effect}
\resizebox{\textwidth}{!}{%
\begin{tabular}{llllrrl}
\toprule
Category & Metric & Direction & PAM60K & Tune60K & $\Delta$ [\%] & Note \\
\midrule
Semantic Segmentation & mIoU & $\uparrow$ & 0.3310 & 0.3115 & -5.90 & PAM better \\
Depth & RMSE & $\downarrow$ & 27.77 & 33.02 & +18.89 & PAM better \\
Edge & L1 Error & $\downarrow$ & 0.04493 & 0.04561 & +1.50 & PAM better \\
Object Preservation & F1 Score & $\uparrow$ & 0.1071 & 0.0889 & -17.04 & PAM better \\
Reality & CLIP-CMMD & $\downarrow$ & 0.1794 & 0.1738 & -3.10 & Tune slightly better \\
Diversity & $1-\mathrm{MS\text{-}SSIM}$ & $\uparrow$ & 0.8480 & 0.8497 & +0.19 & Difference very small \\
Text Alignment & R-Precision@1 & $\uparrow$ & 0.3258 & 0.3563 & +9.37 & Tune better \\
\bottomrule
\end{tabular}}
\end{table}

The cost is a small decrease in CLIP-CMMD and R-Precision@1, while diversity changes negligibly. This is consistent with the design: hard local selection stabilizes structure but slightly constrains style freedom and text-driven appearance variation. Since the objective is annotation-preserving augmentation, this trade-off is acceptable when structural metrics are prioritized.

\begin{figure}[tbp]
\centering
\includegraphics[width=0.95\linewidth]{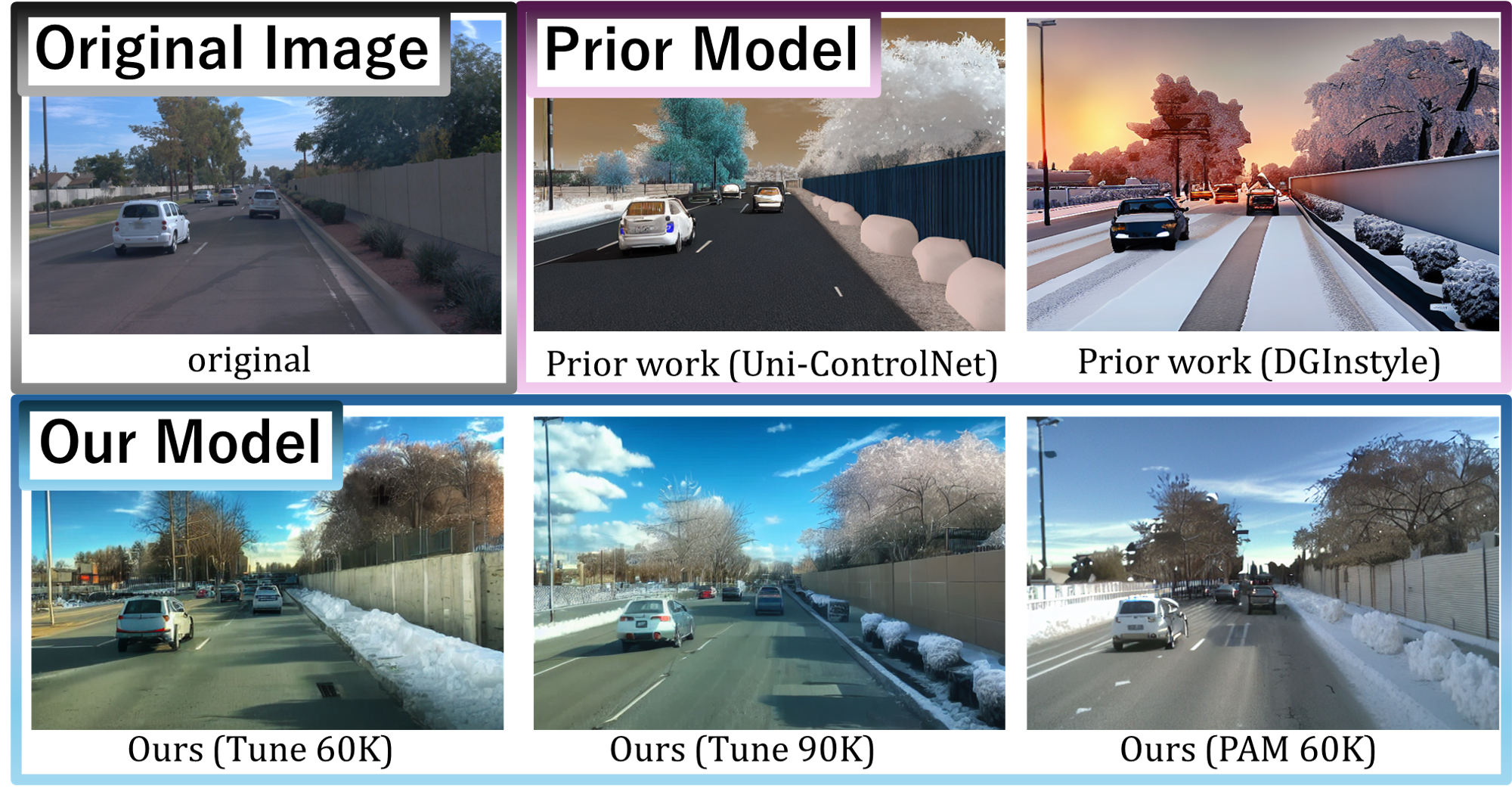}
\caption{Qualitative comparison around \pamshort{}. The models change the original appearance toward a snowy road condition, but distant structural consistency differs.}
\label{fig:qualitative_pam}
\end{figure}

\begin{figure}[tbp]
\centering
\includegraphics[width=0.95\linewidth]{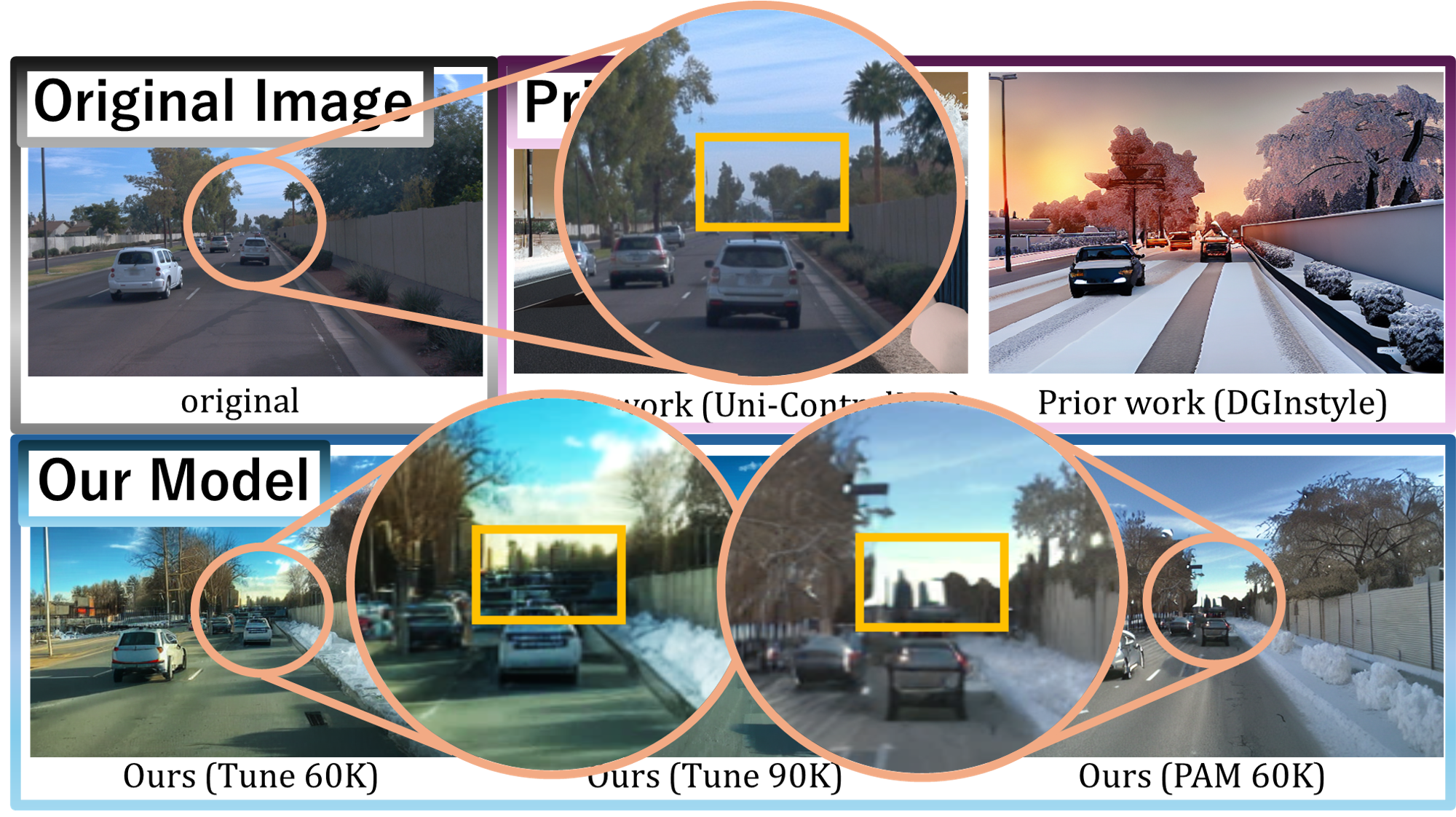}
\caption{Zoomed example of distant structure preservation. Tune60K weakens road continuity and distorts distant trees, while PAM60K preserves the distant structure more consistently.}
\label{fig:qualitative_pam_zoom}
\end{figure}

\Cref{fig:qualitative_pam} shows qualitative results. Both Tune60K and PAM60K can change the original scene to a snowy appearance. However, the effect of \pamshort{} is clearer in the zoomed comparison in \Cref{fig:qualitative_pam_zoom}. Without \pamshort{}, the distant road can break into a stair-like structure and distant trees may deform. With \pamshort{}, road continuity and distant tree shape are better preserved. This visually supports the idea that selecting locally effective conditions reduces conflicts in distant, structurally delicate regions.

\subsection{Comparison with prior work}

\Cref{tab:prior_comparison} compares PAM60K, Tune60K, and DGInStyle. The column ``best ours'' selects the better of PAM60K and Tune60K for each metric. DGInStyle is best for semantic mIoU, diversity, and text alignment. However, the proposed models are best for depth, edge, object preservation, and realism. The largest relative gains over DGInStyle are 51.04\% for edge L1, 35.57\% for object F1, 35.87\% for realism, and 24.35\% for depth RMSE.

\begin{table}[tbp]
\centering
\caption{Comparison with prior work. Positive $\Delta_{\mathrm{Ours\ vs\ DG}}$ means the best proposed model is better than DGInStyle under the metric direction.}
\label{tab:prior_comparison}
\resizebox{\textwidth}{!}{%
\begin{tabular}{lllrrrrrr}
\toprule
Category & Metric & Direction & PAM60K & Tune60K & DGInStyle & Rank 1 & Best ours & $\Delta_{\mathrm{Ours\ vs\ DG}}$ [\%] \\
\midrule
Semantic Segmentation & mIoU & $\uparrow$ & 0.3310 & 0.3115 & 0.3722 & DGInStyle & PAM60K & -11.07 \\
Depth & RMSE & $\downarrow$ & 27.77 & 33.02 & 36.71 & PAM60K & PAM60K & +24.35 \\
Edge & L1 Error & $\downarrow$ & 0.04493 & 0.04561 & 0.09176 & PAM60K & PAM60K & +51.04 \\
Object Preservation & F1 Score & $\uparrow$ & 0.1071 & 0.0889 & 0.0790 & PAM60K & PAM60K & +35.57 \\
Reality & CLIP-CMMD & $\downarrow$ & 0.1794 & 0.1738 & 0.2710 & Tune60K & Tune60K & +35.87 \\
Diversity & $1-\mathrm{MS\text{-}SSIM}$ & $\uparrow$ & 0.8480 & 0.8497 & 0.9240 & DGInStyle & Tune60K & -8.04 \\
Text Alignment & R-Precision@1 & $\uparrow$ & 0.3258 & 0.3563 & 0.3606 & DGInStyle & Tune60K & -1.19 \\
\bottomrule
\end{tabular}}
\end{table}

This comparison clarifies the role of the method. DGInStyle is optimized for semantic-mask-conditioned generation; therefore it is strong in semantic mIoU and allows more appearance freedom. The proposed method deliberately adds depth and edge constraints. This can reduce diversity and slightly reduce text alignment, but it substantially improves geometry, contours, object existence, and realism. For high-level driving tasks, those properties matter because labels are often tied to road geometry, sign presence, lane continuity, and object relationships rather than class mask area alone.

\section{Discussion}
\label{sec:discussion}

\subsection{Why pretraining matters}

The pretraining result suggests that Uni-ControlNet contains useful local-condition interpretation priors. In this work the Stable Diffusion U-Net is frozen, so early fine-tuning depends heavily on how well the local adapter maps structural conditions into U-Net control tensors. Scratch training must learn edge, depth, and segmentation interpretation simultaneously from about 150K images. Pretrained initialization already knows how to connect many local controls to a text-to-image diffusion model, so the limited training budget can focus on adapting to autonomous-driving structure. The large object-preservation gain is consistent with this interpretation because object preservation requires class, contour, and depth consistency at once.

The slightly better CLIP-CMMD of the scratch model at 30K is not evidence that scratch training is generally superior. Uni-ControlNet pretraining is based on broad web-scale data and can carry non-driving or non-photorealistic priors. At short fine-tuning, some of these priors may remain, causing occasional cyber-like or anime-like texture. This explains why pretraining strongly improves structure but can very slightly hurt realism early.

\subsection{Scaling and saturation}

The scaling experiment shows that structure preservation and text/diversity metrics are not in a simple unavoidable trade-off. From 0 to 60K, improving local-condition injection also improves diversity and R-Precision@5. This happens because the diffusion backbone remains frozen: the model can learn how to inject structural conditions without destroying the pretrained text-to-image distribution. However, CLIP-CMMD slightly worsens at 90K, suggesting that once structural adaptation is strong, additional training may restrict style freedom. The curves also indicate saturation near 90K, likely due to the limited training scale, $512\times512$ resolution, and SD1.5-style U-Net capacity.

\subsection{Why \pamshort{} improves depth and object preservation}

\pamshort{} improves all structural metrics, but its largest effects are depth RMSE and object F1. Depth is a low-frequency geometric signal. It can be disrupted when high-frequency edge constraints dominate the same local region. By selecting which condition should dominate each local feature location, \pamshort{} can let depth guide road and distance structure while allowing edge or segmentation to guide appropriate regions. Object F1 also benefits because a traffic object must remain present with plausible shape and location. It is not enough for a mask class to be roughly correct; the object must be detectable after generation.

The small realism and text-alignment decrease is interpretable as the cost of hard selection. \pamshort{} intentionally prevents continuous mixing of all conditions at every location. This stabilizes local structure but can slightly limit the freedom of text-driven style changes. A future variant could combine hard selection for unstable regions with soft mixing for stable regions.

\subsection{Suitability for high-level annotation-preserving augmentation}

The method is most suitable for high-level annotations whose correctness depends on structural information: lane-sign relationships, traffic-rule extraction, HD-map topology, traffic knowledge graphs, and driving Graph VQA. For these tasks, geometry, object presence, and contour preservation are essential. The method is not a guarantee that every high-level label can be reused without validation. If a label directly depends on weather, visibility, night/day state, road slipperiness, sign-text readability, or signal state, it may need rechecking after generation. Therefore, a realistic deployment is to use \method{} as a structure-preserving appearance generator, then apply rule recomputation, automatic consistency checks, or targeted relabeling when the high-level label itself depends on the changed appearance.

\subsection{Limitations}

First, this paper evaluates generated-image properties rather than downstream task improvement. The final test should add generated images to training data and measure performance on semantic segmentation, object detection, lane-sign correspondence, traffic-rule extraction, HD-map topology, or Graph VQA. Second, evaluation is centered on Waymo front-camera images, so generalization to other regions, camera intrinsics, side/rear cameras, and multi-view settings remains open. Third, the resolution is $512\times512$ and the backbone is SD1.5-style U-Net, which limits distant small-object and sign-text preservation. Fourth, common-projection metrics depend on the external projectors; they measure consistency through OneFormer, Metric3Dv2, Canny, and Grounding DINO, not human ground truth. Fifth, the conditions do not explicitly encode lane instances, sign OCR, signal state, map priors, or temporal motion. Finally, hard selection in \pamshort{} can restrict feature combination when multiple conditions are genuinely necessary at one location.

\subsection{Future work}

Future work should evaluate downstream performance after augmentation, expand to multiple datasets and camera views, increase resolution, and use larger transformer or DiT-style backbones. Video augmentation with temporal consistency is also important for tracking, planning, and video-based reasoning. The condition set should be extended to lane instance maps, panoptic segmentation, surface normals, map priors, traffic-sign crops, OCR, and signal states. \pamshort{} could be extended to time-aware selection, scale-dependent selection, reliability-weighted gating, or hybrid hard/soft selection. Finally, evaluation should add relation-consistency metrics and high-level labels so that the question changes from ``is the image visually good?'' to ``is it useful training data for high-level driving intelligence?''

\section{Conclusion}
\label{sec:conclusion}

This paper introduced \method, a multi-condition diffusion framework for structure-preserving synthetic data augmentation in autonomous-driving scenes. The framework changes weather and time-of-day appearance while conditioning on semantic segmentation, depth, and edge maps. It reuses Uni-ControlNet-compatible local-control initialization and introduces \pamshort{} to suppress local condition conflict through trainable hard selection. Experiments on 3,048 Waymo front-camera images show that pretraining improves most structure and alignment metrics, that training-step scaling improves structural consistency, and that \pamshort{} improves semantic, depth, edge, and object-preservation metrics at 60K steps. Compared with DGInStyle, the proposed models are weaker in semantic mIoU but stronger in depth consistency, edge consistency, object preservation, and realism.

The central implication is that high-level annotation-preserving augmentation should not be evaluated only by semantic-mask fidelity. Geometry, contours, object existence, realism, diversity, and prompt alignment must be considered jointly. \method{} is therefore a step toward practical generative augmentation for high-level autonomous-driving tasks where annotations are expensive and structure must be retained.

\appendix

\section{Prompt-generation details}
\label{app:prompt_generation}

\subsection{Weather and time estimation with CLIP}

For an input image $I$, CLIP maps the image to an embedding
\begin{equation}
 z_I=f_{\mathrm{image}}(I).
\end{equation}
For a set of candidate text labels $T=\{t_1,\ldots,t_K\}$, the text encoder computes
\begin{equation}
 z_{t_k}=f_{\mathrm{text}}(t_k).
\end{equation}
The cosine similarity is
\begin{equation}
 s_k=\frac{z_I^\top z_{t_k}}{\|z_I\|\|z_{t_k}\|},
\end{equation}
and the chosen label is $k^*=\arg\max_k s_k$. In the implementation, weather and time are estimated independently. Weather candidates are Clear, Cloudy, Rainy, Snowy, and Foggy; time candidates are Day, Twilight, and Night.

\subsection{Caption generation}

Caption generation is shared between training and evaluation. Qwen3-VL-32B-Instruct receives the RGB image, object names extracted from Cityscapes-style classes, and a fixed instruction. The instruction asks for a concise but semantically dense driving-scene description, focusing on listed objects and spatial relations, background, viewpoint, and image quality. It also forbids weather/time words such as rain, snow, fog, clear/sunny, day, night, dawn, and twilight. The generation uses non-thinking mode, maximum new tokens of 96, and deterministic decoding. Reasoning prefixes and any thinking blocks are removed from the output.

\begin{algorithm}[tbp]
\caption{Shared caption generation without explicit weather/time words}
\label{alg:caption_generation}
\begin{algorithmic}[1]
\Require RGB image $I$, object-name list $C$, image-level seed $s$
\Ensure Clean caption $c$
\State Build the fixed Qwen instruction using $C$
\State Set the random seed to $s$
\State Run Qwen3-VL-32B-Instruct on $(I,\mathrm{instruction})$
\State Disable thinking mode
\State Use maximum 96 new tokens
\State Remove reasoning prefixes and any \texttt{<think>} blocks
\State Collapse repeated whitespace
\State \Return cleaned caption $c$
\end{algorithmic}
\end{algorithm}

\subsection{Training prompts}

Training prompts describe the source image rather than creating a counterfactual target. After CLIP estimates source weather and time, the final prompt is
\begin{quote}\small
\texttt{<caption> Image taken in <source-weather> weather at <source-time>.}
\end{quote}
The source subgroup is saved, and the target subgroup is empty.

\begin{algorithm}[tbp]
\caption{Training prompt generation for source-domain images}
\label{alg:training_prompt}
\begin{algorithmic}[1]
\Require Training RGB image $I$, paired or predicted semantic map $M$, base seed $s_0$, image index $i$
\Ensure Training prompt $p_{\mathrm{train}}$
\State Extract train IDs from $M$ and convert them into class names $C$
\State Encode $I$ with the CLIP image encoder
\State Encode weather labels and time labels with the CLIP text encoder
\State Independently compute weather and time scores
\State Obtain $(w_{\mathrm{src}},t_{\mathrm{src}})$ by argmax on each axis
\If{a cached raw caption exists and resume mode is enabled}
  \State Load cached caption $c$
\Else
  \State Set $s\gets s_0+i$
  \State Generate $c$ using \Cref{alg:caption_generation}
\EndIf
\State Append the source-weather/time sentence to $c$
\State Save image path, source subgroup, raw caption, and final training prompt
\State \Return $p_{\mathrm{train}}$
\end{algorithmic}
\end{algorithm}

\subsection{Evaluation prompts}

Evaluation prompts intentionally change both weather and time. If the source subgroup is $(w_{\mathrm{src}},t_{\mathrm{src}})$, then
\begin{equation}
 w_{\mathrm{tgt}}\sim\mathrm{Uniform}(W\setminus\{w_{\mathrm{src}}\}),\qquad
 t_{\mathrm{tgt}}\sim\mathrm{Uniform}(T\setminus\{t_{\mathrm{src}}\}).
\end{equation}
The advanced template uses the adjective and three decorations from the style dictionary. When class names are missing, the class string is replaced by \texttt{typical urban street elements}.

\begin{algorithm}[tbp]
\caption{Evaluation prompt generation for Waymo front images}
\label{alg:evaluation_prompt}
\begin{algorithmic}[1]
\Require Waymo RGB image $I$, optional predicted train-ID map $M$, base seed $s_0$, image index $i$
\Ensure Evaluation prompt $p_{\mathrm{eval}}$
\If{$M$ is available}
  \State Extract class names $C$ from $M$
\Else
  \State Set $C\gets [\ ]$
\EndIf
\State Estimate $(w_{\mathrm{src}},t_{\mathrm{src}})$ with CLIP
\State Sample $w_{\mathrm{tgt}}$ uniformly from $W\setminus\{w_{\mathrm{src}}\}$
\State Sample $t_{\mathrm{tgt}}$ uniformly from $T\setminus\{t_{\mathrm{src}}\}$
\If{a cached raw caption exists and resume mode is enabled}
  \State Load cached caption $c$
\Else
  \State Set $s\gets s_0+i$ and generate $c$ using \Cref{alg:caption_generation}
\EndIf
\State Fetch adjective and decorations from the style dictionary
\State Join class names or use a fallback class string
\State Build the final advanced prompt
\State Save source subgroup, target subgroup, caption, and final prompt
\State \Return $p_{\mathrm{eval}}$
\end{algorithmic}
\end{algorithm}

\section{Model and \pamshort{} implementation details}
\label{app:model_details}

\subsection{Base generator implementation}

The base generator keeps the Uni-ControlNet unified structure for compatibility. The global condition vector is fixed to $g=0\in\mathbb{R}^{768}$. The local conditions are packed into 21 channels according to the seven Uni-ControlNet local-condition slots. Only edge, depth, and segmentation slots are nonzero. The standard diffusion noise-prediction loss is
\begin{equation}
 \mathcal{L}_{\mathrm{diff}}=\mathbb{E}_{x,\epsilon,t}\left[\|\epsilon-\hat\epsilon\|_2^2\right].
\end{equation}
The denoising U-Net, VAE, and text encoder are fixed, and the control branch is updated.

\begin{table}[tbp]
\centering
\caption{Main components of the base generator without \pamshort{}.}
\label{tab:base_generator_details}
\resizebox{\textwidth}{!}{%
\begin{tabular}{p{40mm}p{110mm}}
\toprule
Item & Description \\
\midrule
Base generator & Stable Diffusion 1.5 style latent diffusion. \\
Initialization & Public Uni-ControlNet-compatible unified weights. \\
Local condition representation & Seven-condition-compatible 21-channel tensor; only edge, depth, and segmentation slots are nonzero. \\
Global condition & 768-dimensional zero vector; no image-content global control is supplied. \\
Local feature stages & Four local feature stages are produced and passed to the local control backend. \\
U-Net injection & Control tensors matched to skip and middle U-Net structures are injected into the frozen U-Net. \\
Frozen modules & Denoising U-Net, VAE, and text encoder. \\
Updated modules & Primarily the local control adapter and, when enabled, \pamshort{} parameters. \\
\bottomrule
\end{tabular}}
\end{table}

\begin{algorithm}[tbp]
\caption{Base local control branch without \pamshort{}}
\label{alg:base_control_branch}
\begin{algorithmic}[1]
\Require Uni-ControlNet-compatible local tensor $C^{\mathrm{uni}}$, noisy latent $z_t$, timestep $t$, text context $c$
\Ensure Control tensors $\{\Delta^{(r)}\}_{r=1}^R$
\State $\{H^{(m)}\}_{m=1}^4 \gets E_{\mathrm{loc}}(C^{\mathrm{uni}})$
\State $\{\Delta^{(r)}\}_{r=1}^R \gets B_{\mathrm{loc}}(z_t,t,c,\{H^{(m)}\}_{m=1}^4)$
\State \Return $\{\Delta^{(r)}\}_{r=1}^R$
\end{algorithmic}
\end{algorithm}

\subsection{Condition-specific stems}

\pamshort{} uses independent stems for edge, depth, and segmentation. Each stem receives a three-channel input and outputs a $192\times64\times64$ local feature grid for $512\times512$ input. The stems are initialized from the first part of the base generator's shared local extractor. For the first input convolution, the corresponding 3-channel slice is copied from the 21-channel Uni-ControlNet input. The slices are edge $(0{:}3)$, depth $(15{:}18)$, and segmentation $(18{:}21)$. Later convolution weights are copied from the shared extractor.

\subsection{Tri-attention gate}

For a local position $u$, let
\begin{equation}
 X_u=[f_u^{\mathrm{edge}},f_u^{\mathrm{dep}},f_u^{\mathrm{seg}}]\in\mathbb{R}^{3\times D}.
\end{equation}
After linear projection and layer normalization, the gate computes query, key, value, and context vectors:
\begin{equation}
 Q_u=W_QX_u,
\quad K_u=W_KX_u,
\quad V_u=W_VX_u,
\quad C_u=W_Cq.
\end{equation}
The context is broadcast to the condition tokens and applied multiplicatively:
\begin{equation}
 \tilde K_u=K_u\odot C_u,
\qquad
 \tilde V_u=V_u\odot C_u.
\end{equation}
The tri-attention output is
\begin{equation}
 Y_u=W_O\left(\mathrm{softmax}\left(\frac{Q_u\tilde K_u^\top}{\sqrt d}\right)\tilde V_u\right).
\end{equation}
Residual and feed-forward updates produce $X''_u$, and the condition score is
\begin{equation}
 s_{u,k}=w_g^\top x''_{u,k}+b_g.
\end{equation}

\subsection{Hard selection and backend connection}

The selection weights are
\begin{equation}
 \pi_{u,k}=\frac{\exp(s_{u,k}/\tau)}{\sum_j\exp(s_{u,j}/\tau)},
\qquad
 y_{u,k}=\mathbf{1}\left[k=\arg\max_j s_{u,j}\right],
\end{equation}
with $\tau=1.0$. The STE weight is
\begin{equation}
 w_{u,k}=\mathrm{stopgrad}(y_{u,k}-\pi_{u,k})+\pi_{u,k}.
\end{equation}
The selected feature is $\tilde f_u=\sum_k w_{u,k}f_u^k$. All local features are reassembled into $\tilde F$, optionally passed through two residual blocks, and then forwarded to the same downstream local control backend as the base generator.

\begin{table}[tbp]
\centering
\caption{Main \pamshort{} settings.}
\label{tab:pam_settings}
\begin{tabular}{ll}
\toprule
Item & Setting \\
\midrule
Target conditions & Edge / depth / segmentation \\
Condition slices & $(0{:}3)$, $(15{:}18)$, $(18{:}21)$ in the 21-channel tensor \\
Stem output & $192\times64\times64$ local feature grid per condition \\
Context & Timestep embedding plus pooled text context \\
Score computation & Tri-attention-style condition scorer \\
Selection & Hard argmax in forward pass, softmax gradient in backward pass \\
Temperature & $\tau=1.0$ \\
Enhancement & Two residual blocks enabled \\
Backend & Tail extractor and local control backend shared with base generator \\
\bottomrule
\end{tabular}
\end{table}

\begin{algorithm}[tbp]
\caption{Local control branch with \pamshort{}}
\label{alg:pam_control_branch}
\begin{algorithmic}[1]
\Require Local tensor $C^{\mathrm{uni}}$, noisy latent $z_t$, timestep $t$, text context $c$
\Ensure Control tensors $\{\Delta^{(r)}_{\mathrm{pam}}\}_{r=1}^R$
\State Slice $(C^{\mathrm{edge}},C^{\mathrm{dep}},C^{\mathrm{seg}})$ from $C^{\mathrm{uni}}$
\State $F^{\mathrm{edge}}\gets S_{\mathrm{edge}}(C^{\mathrm{edge}})$
\State $F^{\mathrm{dep}}\gets S_{\mathrm{dep}}(C^{\mathrm{dep}})$
\State $F^{\mathrm{seg}}\gets S_{\mathrm{seg}}(C^{\mathrm{seg}})$
\State $q\gets\psi_t(t)+\psi_c(\mathrm{Pool}(c))$
\For{each local position $u$}
  \State $X_u\gets[f_u^{\mathrm{edge}},f_u^{\mathrm{dep}},f_u^{\mathrm{seg}}]$
  \State $s_u\gets\mathrm{TriAttentionGate}(X_u,q)$
  \State $w_u\gets\mathrm{STEHardSelect}(s_u)$
  \State $\tilde f_u\gets\sum_k w_{u,k}f_u^k$
\EndFor
\State Reassemble $\{\tilde f_u\}$ into $\tilde F$
\State Generate four local feature levels with the shared tail extractor
\State $\{\Delta^{(r)}_{\mathrm{pam}}\}_{r=1}^R\gets B_{\mathrm{loc}}(z_t,t,c,\{H^{(m)}_{\mathrm{pam}}\}_{m=1}^4)$
\State \Return $\{\Delta^{(r)}_{\mathrm{pam}}\}_{r=1}^R$
\end{algorithmic}
\end{algorithm}

\section{Evaluation details}
\label{app:evaluation_details}

\subsection{Common projection}

Let the paired original and generated image sets be
\begin{equation}
 \mathcal{X}=\{x_i\}_{i=1}^N,
\qquad
 \mathcal{Y}=\{y_i\}_{i=1}^N,
\qquad y_i=F(x_i).
\end{equation}
Let $P_{\mathrm{seg}}$, $P_{\mathrm{dep}}$, $P_{\mathrm{edge}}$, and $P_{\mathrm{obj}}$ be the semantic, depth, edge, and object projectors. For image pair $(x_i,y_i)$,
\begin{equation}
 G_i^{(m)}=P_m(x_i),
\qquad
 \hat G_i^{(m)}=P_m(y_i).
\end{equation}
The score is computed by a metric-specific comparison function $\Psi_m$ over paired projections:
\begin{equation}
 S_m=\Psi_m\left(\{(G_i^{(m)},\hat G_i^{(m)})\}_{i=1}^N\right).
\end{equation}
This evaluates pairwise structural consistency through a common observer, not absolute projector accuracy.

\begin{algorithm}[tbp]
\caption{Common-projection structure evaluation}
\label{alg:common_projection}
\begin{algorithmic}[1]
\Require Original images $\mathcal{X}$ and generated images $\mathcal{Y}$
\Ensure Semantic mIoU, depth RMSE, edge L1, and object F1
\For{$i=1$ to $N$}
  \State Predict semantic maps on $x_i$ and $y_i$ using OneFormer
  \State Predict relative depth maps on $x_i$ and $y_i$ using Metric3Dv2
  \State Extract edge maps on $x_i$ and $y_i$ using Canny
  \State Detect target objects on $x_i$ and $y_i$ using Grounding DINO
  \State Treat all predictions from $x_i$ as pseudo ground truth
  \State Accumulate pairwise matches and errors
\EndFor
\State Aggregate image-level scores into dataset-level scores
\State \Return the four structure-preservation scores
\end{algorithmic}
\end{algorithm}

\subsection{Semantic mIoU}

Let $S_i^{\mathrm{src}}$ and $S_i^{\mathrm{gen}}$ be the OneFormer label maps of original and generated images. For the valid class set
\begin{equation}
 C_i^{\mathrm{valid}}=\left\{c\in C\;\middle|\; |[S_i^{\mathrm{src}}=c]\cup[S_i^{\mathrm{gen}}=c]|>0\right\},
\end{equation}
the image-level IoU is
\begin{equation}
 \mathrm{IoU}_{i,c}=\frac{|[S_i^{\mathrm{src}}=c]\cap[S_i^{\mathrm{gen}}=c]|}{|[S_i^{\mathrm{src}}=c]\cup[S_i^{\mathrm{gen}}=c]|}.
\end{equation}
The dataset mIoU is
\begin{equation}
 \mathrm{mIoU}=\frac{1}{N}\sum_{i=1}^N\frac{1}{|C_i^{\mathrm{valid}}|}\sum_{c\in C_i^{\mathrm{valid}}}\mathrm{IoU}_{i,c}.
\end{equation}

\subsection{Depth RMSE}

Let $D_i^{\mathrm{src}}$ and $D_i^{\mathrm{gen}}$ be Metric3Dv2 relative depth maps. For valid depth pixels $\Omega_i^{\mathrm{dep}}$, the depth RMSE is
\begin{equation}
 \mathrm{RMSE}_{\mathrm{dep}}=\sqrt{\frac{\sum_{i=1}^N\sum_{u\in\Omega_i^{\mathrm{dep}}}(D_i^{\mathrm{src}}(u)-D_i^{\mathrm{gen}}(u))^2}{\sum_{i=1}^N|\Omega_i^{\mathrm{dep}}|}}.
\end{equation}
It is interpreted as relative-geometry consistency, not absolute distance accuracy.

\subsection{Edge L1}

Edge maps $E_i^{\mathrm{src}}$ and $E_i^{\mathrm{gen}}$ are extracted by Canny. The edge metric uses a mask consisting of the lower half of the image and traffic-sign regions. This avoids overemphasizing sky and distant texture while retaining driving-relevant road and sign structure. Let
\begin{equation}
 M_i^{\mathrm{edge}}=M_i^{\mathrm{low}}\vee M_i^{\mathrm{sign}}.
\end{equation}
The masked edge L1 error is
\begin{equation}
 \mathrm{L1}_{\mathrm{edge}}=\frac{\sum_{i=1}^N\sum_u M_i^{\mathrm{edge}}(u)|E_i^{\mathrm{src}}(u)-E_i^{\mathrm{gen}}(u)|}{\sum_{i=1}^N\sum_u M_i^{\mathrm{edge}}(u)}.
\end{equation}

\begin{algorithm}[tbp]
\caption{Masked edge evaluation}
\label{alg:edge_evaluation}
\begin{algorithmic}[1]
\Require Original image $x_i$ and generated image $y_i$
\Ensure Masked edge L1 error
\State Extract Canny edge maps $E_i^{\mathrm{src}}$ and $E_i^{\mathrm{gen}}$
\State Build the lower-half mask $M_i^{\mathrm{low}}$
\State Detect traffic-sign boxes on $x_i$ and convert them into an ROI mask $M_i^{\mathrm{sign}}$
\State Set $M_i^{\mathrm{edge}}\gets M_i^{\mathrm{low}}\vee M_i^{\mathrm{sign}}$
\State Compute and normalize the masked absolute difference
\State \Return pairwise masked L1 error
\end{algorithmic}
\end{algorithm}

\subsection{Object-preservation F1}

Grounding DINO detects traffic-related classes:
\begin{quote}\small
car, truck, bus, motorcycle, bicycle, person, pedestrian, traffic light, traffic sign, stop sign, speed limit sign, crosswalk sign, construction sign, traffic cone.
\end{quote}
For each class $c$, original detections and generated detections are matched one-to-one if their IoU is at least 0.5. Let $\mathrm{TP}_c$, $\mathrm{FP}_c$, and $\mathrm{FN}_c$ be the resulting counts. The class F1 is
\begin{equation}
 \mathrm{F1}_c=\frac{2\mathrm{TP}_c}{2\mathrm{TP}_c+\mathrm{FP}_c+\mathrm{FN}_c},
\end{equation}
and the reported object score is the macro average over target classes.

\begin{algorithm}[tbp]
\caption{Object-preservation F1 for traffic-object classes}
\label{alg:object_f1}
\begin{algorithmic}[1]
\Require Paired images $\{(x_i,y_i)\}_{i=1}^N$, target class set $C_{\mathrm{obj}}$
\Ensure Macro-averaged object F1
\For{each class $c\in C_{\mathrm{obj}}$}
  \State Initialize $\mathrm{TP}_c$, $\mathrm{FP}_c$, and $\mathrm{FN}_c$ to zero
  \For{$i=1$ to $N$}
    \State Detect class-$c$ boxes on $x_i$ and $y_i$ with Grounding DINO
    \State Match boxes one-to-one using IoU threshold 0.5
    \State Accumulate matched, generated-only, and original-only boxes
  \EndFor
  \State Compute $\mathrm{F1}_c$
\EndFor
\State Average over all classes in $C_{\mathrm{obj}}$
\State \Return macro-averaged object F1
\end{algorithmic}
\end{algorithm}

\subsection{CLIP-CMMD}

Let $z_i=\phi_{\mathrm{img}}(x_i)$ and $\tilde z_j=\phi_{\mathrm{img}}(y_j)$ be CLIP image embeddings. CLIP-CMMD is MMD in this embedding space:
\begin{align}
\mathrm{CMMD}^2(\mathcal{X},\mathcal{Y})
&=\frac{1}{N(N-1)}\sum_{i\ne j}k(z_i,z_j)
+\frac{1}{M(M-1)}\sum_{i\ne j}k(\tilde z_i,\tilde z_j)\\
&\quad -\frac{2}{NM}\sum_{i=1}^{N}\sum_{j=1}^{M}k(z_i,\tilde z_j).
\end{align}
Smaller is better, and zero means the two empirical distributions match in kernel mean embedding.

\begin{algorithm}[tbp]
\caption{CLIP-CMMD evaluation}
\label{alg:cmmd}
\begin{algorithmic}[1]
\Require Original image set $\mathcal{X}$ and generated image set $\mathcal{Y}$
\Ensure CLIP-CMMD score
\State Extract CLIP image embeddings for all images in $\mathcal{X}$
\State Extract CLIP image embeddings for all images in $\mathcal{Y}$
\State Build empirical kernel matrices
\State Compute empirical MMD between embedding sets
\State \Return CMMD score
\end{algorithmic}
\end{algorithm}

\subsection{Diversity metrics}

For generated image set $\mathcal{Y}=\{y_i\}_{i=1}^N$, sample a pair set $\mathcal{P}\subset\{(i,j):1\le i<j\le N\}$. LPIPS uses deep feature distances with AlexNet features:
\begin{equation}
 d_{\mathrm{LPIPS}}(y_i,y_j)=\sum_l\frac{1}{H_lW_l}\left\|w_l\odot\left(\hat\phi_l(y_i)-\hat\phi_l(y_j)\right)\right\|_2^2.
\end{equation}
The average LPIPS is the mean over $\mathcal{P}$. MS-SSIM is converted to dissimilarity as
\begin{equation}
 d_{\mathrm{MS\text{-}SSIM}}(y_i,y_j)=1-\mathrm{MS\text{-}SSIM}(y_i,y_j).
\end{equation}
Both diversity metrics are larger-is-better.

\begin{algorithm}[tbp]
\caption{Diversity evaluation with AlexNet-LPIPS and $1-\mathrm{MS\text{-}SSIM}$}
\label{alg:diversity}
\begin{algorithmic}[1]
\Require Generated image set $\mathcal{Y}$
\Ensure Average LPIPS and average $1-\mathrm{MS\text{-}SSIM}$
\State Randomly sample image pairs $\mathcal{P}$ from $\mathcal{Y}$
\For{each pair $(y_i,y_j)\in\mathcal{P}$}
  \State Compute AlexNet-LPIPS distance
  \State Compute $1-\mathrm{MS\text{-}SSIM}$
\EndFor
\State Average both scores over all sampled pairs
\State \Return diversity scores
\end{algorithmic}
\end{algorithm}

\subsection{CLIP-R-Precision}

For each generated image $y_i$, let $t_i^+$ be its matched prompt and let $\{t_{i,1}^-,\ldots,t_{i,99}^-\}$ be mismatched prompts. The candidate set has 100 prompts. CLIP similarity is
\begin{equation}
 s_{i,j}=\cos\left(\phi_{\mathrm{img}}(y_i),\phi_{\mathrm{text}}(t_{i,j})\right).
\end{equation}
Let $r_i$ be the rank of the matched prompt after sorting candidates by similarity. Then
\begin{equation}
 \mathrm{R\text{-}Precision@K}=\frac{1}{N}\sum_{i=1}^N\mathbf{1}[r_i\le K].
\end{equation}
For 100 candidates, random ranking gives an expectation of about $K/100$.

\begin{algorithm}[tbp]
\caption{CLIP-R-Precision evaluation}
\label{alg:rprecision}
\begin{algorithmic}[1]
\Require Generated images $\{y_i\}_{i=1}^N$, matched prompts $\{t_i^+\}_{i=1}^N$, mismatch prompt pool, depth $K$
\Ensure R-Precision@K
\For{$i=1$ to $N$}
  \State Construct 100 candidate prompts with one matched prompt and 99 mismatched prompts
  \State Encode $y_i$ with the CLIP image encoder
  \State Encode all candidates with the CLIP text encoder
  \State Rank candidates by image-text similarity
  \State Record a hit if the matched prompt is within top-$K$
\EndFor
\State Average hit indicators over all generated images
\State \Return R-Precision@K
\end{algorithmic}
\end{algorithm}

\bibliographystyle{plain}
\bibliography{thesis}

\end{document}